%% file: arXiv.tex
\documentclass[11pt]{article}

\usepackage[final]{acl}

\usepackage{times}
\usepackage{latexsym}

\usepackage[T1]{fontenc}

\usepackage[utf8]{inputenc}

\usepackage{microtype}

\usepackage{inconsolata}

\usepackage{graphicx}

\usepackage{enumitem}
\usepackage{amsmath}
\usepackage{bbm}
\usepackage{tabularx}
\usepackage{booktabs} 
\usepackage{multirow} 
\usepackage{amssymb}
\usepackage{mathabx}
\usepackage{tcolorbox}
\usepackage{makecell}

\newcommand{\commentZH}[1]{#1}
\newcommand{\hashline}{\#\#\#\#\#\#\#\#\#\#\#\#\#\#\#\#}

\title{Knowledge-to-Verification: Exploring RLVR for LLMs in Knowledge-Intensive Domains}

\author{
 \textbf{Zhonghang Yuan}\textsuperscript{\textmd{1}}\footnotemark[1],
 \textbf{Zhefan Wang}\textsuperscript{\textmd{1}}\footnotemark[1],
 \textbf{Fang Hu}\textsuperscript{\textmd{2}}\footnotemark[1],
 \textbf{Zihong Chen}\textsuperscript{\textmd{1}},
 \textbf{Jinzhe Li}\textsuperscript{\textmd{1}},
 \textbf{Gang Li}\textsuperscript{\textmd{1}},
 \\
 \textbf{Jie Ying}\textsuperscript{1},
 \textbf{Huanjun Kong}\textsuperscript{1},
 \textbf{Songyang Zhang}\textsuperscript{1},
 \textbf{Nanqing Dong}\textsuperscript{1,2}\footnotemark[2]
\\
 \textsuperscript{1}Shanghai Artificial Intelligence Laboratory
\\
 \textsuperscript{2}Shanghai Innovation Institute
\\
}

\begin{document}
\maketitle

\renewcommand{\thefootnote}{\fnsymbol{footnote}}
\footnotetext[1]{Equal contribution. }
\footnotetext[2]{Corresponding author.}
\renewcommand{\thefootnote}{\arabic{footnote}}

\begin{abstract}
Reinforcement learning with verifiable rewards (RLVR) has demonstrated promising potential to enhance the reasoning capabilities of large language models (LLMs) in domains such as mathematics and coding. However, its applications on knowledge-intensive domains have not been effectively explored due to the scarcity of high-quality verifiable data. 
Furthermore, current RLVR focuses solely on the correctness of final answers, leading to the limitations of flawed reasoning and sparse reward signals. In this work, we propose \textbf{K}nowledge-\textbf{t}o-\textbf{V}erification (\textbf{K2V}), a framework that extends RLVR to knowledge-intensive domains through automated verifiable data synthesis, while enabling verification of the LLM's reasoning process. 
Extensive experiments demonstrate that K2V enhances the reasoning of LLM in knowledge-intensive domains without significantly compromising the model's general capabilities.
This study also suggests that integrating automated data synthesis with reasoning verification is a promising direction to enhance model capabilities in these broader domains. Code is available at~\url{https://github.com/SeedScientist/K2V}.
\end{abstract}

\section{Introduction}

Recent large language models (LLMs), such as OpenAI-o1~\citep{jaech2024openai}, DeepSeek-R1~\citep{guo2025deepseek}, and Qwen3~\citep{yang2025qwen3}, have demonstrated remarkable progress in reasoning. Central to this progress is reinforcement learning with verifiable rewards (RLVR)~\citep{shen2025satori, peng2025rewarding, stojanovski2025reasoning}, which drives the model to self-explore during training by comparing its outputs against a verifiable ground truth, thereby enhancing its capacity for complex problem-solving.

However, current RLVR methods are confined to mathematical reasoning~\citep{zeng2025simplerl, liu2025understanding} and coding tasks~\citep{he2025skywork, luo2025deepcoder, cui2025process}, lacking the transferability to knowledge-intensive domains (\emph{e.g.}~agriculture, law, and medicine), which heavily rely on specialized knowledge. 
This narrow focus can be attributed to two main reasons: \textit{unverifiable answers} and \textit{limited data}. Firstly, for mathematics, the correctness of an LLM's response can be directly validated by a rule-based verifier~\citep{hu2025open}, and for coding, unit tests can be directly executed on model-generated code in a sandboxed environment \citep{luo2025deepcoder}. However, in knowledge-intensive domains, the answers are typically in the form of open-ended text, which cannot be automatically validated. 
Secondly, in the domains of mathematics and coding, a vast amount of verifiable data can be acquired from the internet and textbooks~\citep{ma2025general}, and synthesizing such data is also relatively straightforward~\citep{yang2025qwen3}. 
In contrast, data collected for knowledge-intensive domains are commonly unverifiable and low quality text. Knowledge-intensive domains also lack effective data synthesis methods, relying on costly, expert-level manual annotation~\citep{dubois2023alpacafarm}.

Moreover, current RLVR methods suffer from two inherent limitations. The first one is \textit{flawed reasoning}. The traditional RLVR focus solely on the correctness of the final answer, ignoring the validity of the reasoning process. This reward mechanism can lead LLM to exhibit issues such as linguistic incoherence~\citep{guo2025deepseek} and unfaithful reasoning~\citep{chen2025reasoning}. The second one is \textit{sparse rewards}. Awarding a binary reward based only on the final answer creates an overly sparse reward signal, which increases the variance of the policy gradient estimate, introduces noise to the training, and results in unstable learning and slower convergence~\citep{su2025klear}.


To address these problems, we propose \textbf{K}nowledge-\textbf{t}o-\textbf{V}erification (\textbf{K2V}), which automatically synthesizes verifiable question-answering (QA) pairs in knowledge-intensive domains, 
while also enabling the verification of the LLMs' reasoning process. 
The motivation of K2V is that the structured knowledge is easier to verify than unstructured knowledge. 
To this end, we present a fill-blank style verification, which first organizes knowledge from the corpora into a knowledge graph (KG)~\citep{hogan2021knowledge}, and then transforms the conventional knowledge graph completion (KGC) task~\citep{ji2021survey, yao2025exploring} into fill-blank style QA pairs. This enables the efficient synthesis of large-scale verifiable QA pairs.
Furthermore, directly verifying the correctness of an LLM's reasoning process is challenging. However, following the principle of \emph{Divide and Conquer}~\citep{cormen2022introduction}, this task can be decomposed into multiple binary-verifiable subtasks.
Specifically, we introduce a checklist-style verification method that generates a checklist for each QA pair. This checklist consists of multiple verifiable subtasks that describe the criteria for a correct reasoning process. Each subtask can be answered with a simple yes or no.
Finally, we propose an answer-gated reward mechanism. This design ensures that the reasoning reward is awarded only when the final answer is correct, thereby anchoring the model's logical consistency to factual accuracy and preventing potential reward hacking.

To evaluate the effectiveness and robustness of K2V, extensive experiments were conducted on three representative knowledge-intensive domains: agriculture, law, and medicine. The results based on Qwen2.5~\citep{qwen2025qwen25technicalreport} and Llama3~\citep{dubey2024llama} backbones show that K2V can enhance the reasoning of LLMs without significantly compromising their general abilities, and in general outperforms the existing baselines that can synthesize verifiable data for knowledge-intensive domains. Ablation studies further suggest that the proposed verification and rewarding designs are simple yet effective, which might be of interest to the broad RLVR community. 


Our contributions are summarized as follows: 
\begin{itemize}[itemsep=0pt, topsep=0pt, parsep=0pt]
    \item We present K2V, a scalable framework that explores RLVR in knowledge-intensive domains. 
    \item We introduce fill-blank style verification, which is designed to synthesize large-scale verifiable QA pairs, and checklist-style verification, which aims to verify the model's reasoning process. 
    \item We integrate answer and reasoning rewards through an answer-gated reward mechanism, which ensures factual correctness and prevents potential reward hacking.
    \item We conduct extensive experiments to demonstrate that K2V enhances model's reasoning capabilities in knowledge-intensive domains without significantly compromising general capabilities.
\end{itemize}

\section{Related Work}
\label{sec:related}
\noindent\textbf{Reinforcement Learning with Verifiable Rewards.}
Unlike conventional reinforcement learning from human feedback (RLHF) that relies on scalar reward models~\citep{ouyang2022training}, RLVR aims to enhance the model's reasoning abilities by computing rewards for tasks that can be automatically verified~\citep{guo2025deepseek}, \emph{e.g.}~mathematics~\citep{zeng2025simplerl} and coding~\citep{he2025skywork}. Thus, for reasoning tasks in the knowledge-intensive domains that can not be automatically verified, existing RLVR studies can not be directly applied. Even though VeriFree~\citep{zhou2025reinforcing} attempts to use the model's internal probability distribution as a reward signal, it still rely on the availability of verifiable data. Furthermore, existing RLVR studies overlook the quality of reasoning process~\citep{chen2025reasoning}, which may lead to flawed reasoning, \emph{e.g.}~studies such as Light-R1~\citep{wen2025light} focus primarily on the correctness of final answers. In this study, we aim to bridge the gap between verifiable data and knowledge-intensive tasks.

\noindent \textbf{Data Synthesis.}
There have been recent studies that leverage data synthesis to improve supervised fine-tuning (SFT) performanc. Liquid~\citep{lee2023liquid}, Synthetic Data RL~\citep{guo2025synthetic}, and BDS~\citep{dedhia2025bottom}  synthesize verifiable QA pairs by extracting key information from the corpus, but fails to establish associations across different tasks. Genie~\citep{yehudai2024genie} and Evol-Instruct~\citep{xu2024wizardlm} primarily focus on open-ended text, which can not be verified. Open-R1~\citep{openr1} and DeepScaleR~\citep{deepscaler2025} focus on synthesizing verifiable data in mathematics or coding domains, but struggling to integrate specialized expertise from knowledge-intensive domains to generate verifiable data. We perform both quantitative and qualitative comparison between K2V and existing RLVR methods that can be applied to knowledge driven tasks in Section~\ref{Main Results} and Appendix~\ref{Case Study}.

\begin{figure*}[t]
    \begin{center}
        \includegraphics[width=\linewidth]{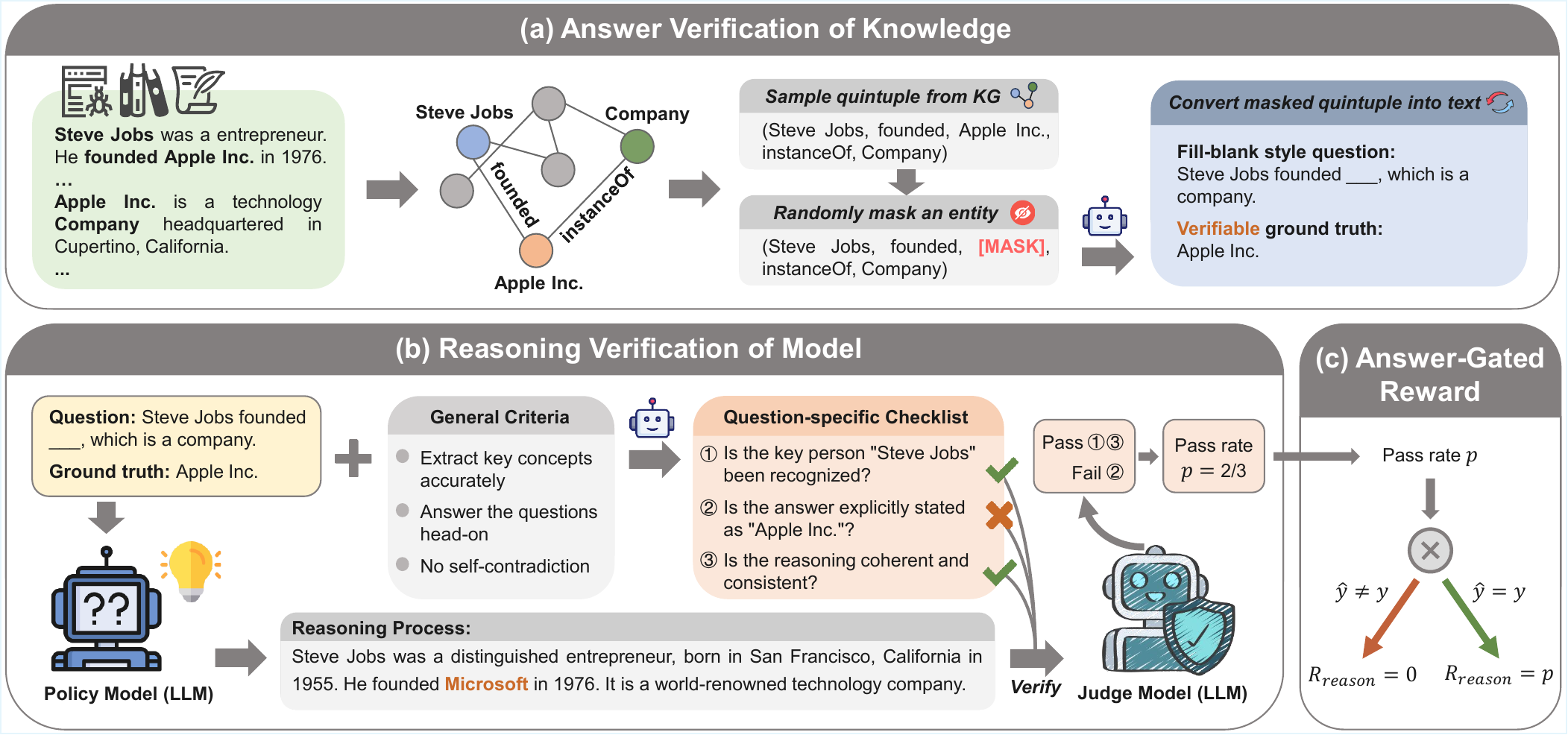}
    \end{center}
    \caption{An overview of K2V. \textbf{(a)} K2V begins by constructing a KG from unstructured corpora. It then samples quintuples from the KG and randomly masks an entity. This masked quintuple is then converted into a fill-blank style question, where the name of the masked entity serves as the verifiable ground truth. \textbf{(b)} Given a QA pair, the Policy Model generates a reasoning process. To verify this reasoning process, K2V first creates a question-specific checklist by instantiating a set of general criteria. The Judge Model then verifies the reasoning process against each item in the checklist. The pass rate on the checklist serves as a dense reward signal. \textbf{(c)} Reasoning reward is awarded only when the predicted answer $\hat{y}$ matches the ground truth $y$. Otherwise, the reasoning reward is 0 regardless of how logical the reasoning process may appear.}
    \label{fig:overview}
\end{figure*}

\section{Knowledge-to-Verification}

In this section, we first present fill-blank style verification (Section \ref{Answer Verification of Knowledge}), \commentZH{which enables the synthesis of verifiable QA pairs.} Next, we introduce a checklist-style verification (Section \ref{Reasoning Verification of Model}), \commentZH{which not only validates the model's reasoning process but also provides dense reward signals.} Finally, we discuss the answer-gated reward mechanism (Section \ref{Answer-Gated Reward Mechanism}), \commentZH{which ensures factual correctness and prevents potential reward hacking.} An overview of K2V is presented in Figure~\ref{fig:overview}.

\subsection{Answer Verification of Knowledge}
\label{Answer Verification of Knowledge}

\commentZH{\noindent\textbf{Knowledge Graph Completion.} 
A KG is a structured representation of factual knowledge, formally defined as $\mathcal{G} = \{\mathcal{E}, \mathcal{R}, \mathcal{T}\}$, where $\mathcal{E}$ is the set of entities, $\mathcal{R}$ is the set of relations, and $\mathcal{T} \subseteq \mathcal{E} \times \mathcal{R} \times \mathcal{E}$ represents the set of triples. Each triple $(h, r, t) \in \mathcal{T}$ consists of a head entity $h$, a relation $r$, and a tail entity $t$. 
The goal of conventional KGC is entity prediction. Given a triple with a missing entity, formally expressed as $(h, r, ?)$ for a missing tail or $(?, r, t)$ for a missing head, the task is to predict the missing entity from the set of entities $\mathcal{E}$. }

\noindent \textbf{Fill-blank Style Verification.}
Following the principles of KGC, we reformulate conventional KGC tasks as fill-blank style QA pairs to synthesize verifiable data in knowledge-intensive domains.
Meanwhile, to ensure the synthesized questions possess sufficient context for complex reasoning, K2V operates not on individual triples, but on quintuples. A quintuple $\sigma$ is defined as:
\begin{equation}
\label{eq:quintuple}
\sigma = (e_1, r_1, e_2, r_2, e_3)
\end{equation}
Where $e_1$, $e_2$, $e_3$ denote entities and $r_1$, $r_2$ denote relations.

To generate a QA pair, K2V first randomly masks one of the three entities (\emph{i.e.},~$e_1$, $e_2$, or $e_3$) to create a masked quintuple $\sigma_{\text{masked}}$, e.g., $(e_1, r_1, \texttt{[MASK]}, r_2, e_3)$. K2V then employs an LLM, denoted as $\mathcal{F}_{\text{text}}$, to convert the masked quintuple into a fill-blank style question $Q_{\text{blank}}$,
\begin{equation}
\label{eq:textualization function}
Q_{\text{blank}} = \mathcal{F}_{\text{text}}(\sigma_{\text{masked}})
\end{equation}
where the ground truth for $Q_{\text{blank}}$ is the name of the masked entity, \commentZH{which can be automatically verified by a rule-based validator.} The prompt of $\mathcal{F}_{\text{text}}$ is shown in Appendix~\ref{Prompt of f_text}.

\noindent \textbf{QA Pairs Synthesis.}
To synthesize verifiable QA pairs, K2V first utilizes the GraphGen~\citep{chen2025graphgen} to construct a KG. Specifically, K2V employs an LLM to perform named entity recognition (NER)~\citep{keraghel2024survey} and relation extraction (RE)~\citep{zhao2024comprehensive} on unstructured corpora (see Appendix~\ref{Prompt for NER and RE} for the prompt), then links the entities and relations to construct a KG.

After the KG is constructed, K2V directly samples quintuples from the KG. For each sampled quintuple $\sigma$ (see Equation~\ref{eq:quintuple}), K2V randomly selects an entity from $\{e_1, e_2, e_3\}$ to be masked, and then uses $\mathcal{F}_{\text{text}}$ (see Equation~\ref{eq:textualization function}) to convert the masked quintuple $\sigma_{\text{masked}}$ into a sentence. This sentence is a fill-blank style question,  with the ground truth being the name of the masked entity.

K2V operates entirely from scratch, requiring no human annotation or seed datasets~\citep{wang2023self, yu2025cot}. This means K2V can be scalable to large-scale unstructured corpora.

\subsection{Reasoning Verification of Model}
\label{Reasoning Verification of Model}

Given an input question $x$ and a policy model $\pi_\theta$ parameterized by $\theta$, a reasoning process $z$ and a response $\hat{y}$ are sampled from the policy, denoted by $z, \hat{y} \sim \pi_\theta(\cdot | x)$. The reasoning process $z$ is typically a lengthy, open-ended text. Due to the lack of evaluation criteria, directly verifying the correctness of $z$ is challenging. Motivated by this, we propose checklist-style verification.

\noindent \textbf{Checklist-Style Verification.}
For each question $x$, we establish a checklist, which is formally represented as a set of $k$ verifiable criteria:
\begin{equation}
\label{eq:checklist}
C^x = \left \{ c_1, c_2, \dots, c_k \right \}
\end{equation}
where each criterion $c_i$ is a binary-verifiable criterion that assesses a desirable property of a reasoning process. These criteria evaluate the policy's reasoning process from different perspectives, and collectively, they form a checklist $C^x$ for a comprehensive assessment of reasoning quality. Most importantly, \textbf{the checklist $C^x$ is question-specific:} a unique checklist is generated for each question $x$. This enables a tailored evaluation of reasoning quality for different questions.

To perform the verification, K2V employs a generative LLM as the judge model, denoted as $J$. The judge model assesses the reasoning process $z$ against each criterion $c_i$ in the checklist $C^x$ one by one (see Appendix~\ref{Prompt of Judge Model} for the prompt). For each pair $(z, c_i)$, the judge model outputs a binary score $v_i \in \{0, 1\}$, where $v_i=1$ indicates that $z$ satisfies the criterion $c_i$, and $v_i=0$ otherwise. We can define this verification process as:
\begin{equation}
\label{eq:judge_model}
v_i = J(z, c_i)
\end{equation}




Once the judge model evaluates all criteria, K2V aggregates these binary scores to compute a pass rate $p\in[0, 1]$, representing the proportion of criteria that the reasoning process $z$ successfully meets:
\begin{equation}
\label{eq:pass rate}
p = \frac{1}{k} \sum_{i=1}^{k} v_i
\end{equation}
where $v_i$ is computed by Equation~\ref{eq:judge_model}. This approach decomposes the intractable reasoning verification task into a series of binary-verifiable subtasks. The pass rate $p$ can serve as a dense reward signal.

\begin{table*}[th]
  \caption{Overall performance on three different knowledge-intensive domains. To evaluate the model's performance more comprehensively in the agricultural domain, we select agriculture-related subsets from CMMLU and MMLU. A similar evaluation strategy is applied to the legal and medical domains. \commentZH{We use the base version for the Qwen backbone and the instruction-tuned version for the Llama backbone.} We \textbf{bold} the best result and \underline{underline} the suboptimal one. Avg denotes the average accuracy of a model in a specific domain.}
  \centering
  \resizebox{\textwidth}{!}{
    \large
    \begin{tabular}{l|cccc|cccc|cccc}
      \toprule[0.6pt] 
      & \multicolumn{4}{c|}{\textbf{Agriculture}} & 
      \multicolumn{4}{c|}{\textbf{Law}} &
      \multicolumn{4}{c}{\textbf{Medicine}} \\
      \multirow{-2}{*}{\textbf{Model}} &  
      \textbf{SeedBench} & \textbf{CMMLU} & \textbf{MMLU} & \textbf{Avg} &
      \textbf{LawBench} & \textbf{CMMLU} & \textbf{MMLU} & \textbf{Avg} & \textbf{MedQA} & \textbf{CMMLU} & \textbf{MMLU} & \textbf{Avg}
      \\
      \midrule[0.6pt] 
       \multicolumn{13}{c}{\textbf{Qwen-2.5-3B (Backbone)}} \\
      \midrule[0.6pt] 
      Qwen2.5-3B-Instruct & 45.67 & 61.23 & 73.94 & 60.28 & 42.39 & 62.82 & \textbf{63.10} & 56.10 & 73.01 & 61.22 & 69.67 & 67.97 \\
      \midrule[0.6pt] 
      Liquid-3B-Qwen & 53.60 & 60.22 & 68.07 & 60.63 & 32.95 & 52.98 & 58.78 & 48.24 & 68.89 & 58.11 & 63.44 & 63.48 \\
      Genie-3B-Qwen & 58.09 & 59.59 & 73.34 & 63.67 & 36.94 & 60.45 & 59.71 & 52.37 & 73.02 & 61.69 & 68.50 & 67.74 \\
      SDR-3B-Qwen & \underline{59.06} & 63.82 & \underline{73.99} & \underline{65.62} & \underline{37.85} & \underline{63.37} & 61.02 & \underline{54.08} & \underline{73.67} & \underline{64.82} & \textbf{70.89} & \underline{69.79} \\
      BDS-3B-Qwen & 54.71 & \underline{65.33} & 70.83 & 63.62 & 32.44 & 68.55 & 60.82 & 53.94 & 72.94 & 64.54 & 62.20 & 66.56 \\
      \midrule[0.6pt]
      \textbf{K2V-3B-Qwen} & \textbf{62.82} & \textbf{66.82} & \textbf{75.40} & \textbf{68.34} & \textbf{43.27} & \textbf{71.55} & \underline{62.01} & \textbf{58.94} & \textbf{78.45} & \textbf{67.53} & \underline{70.76} & \textbf{72.24} \\
      \midrule[0.6pt] 
      
      \multicolumn{13}{c}{\textbf{Qwen-2.5-7B (Backbone)}} \\
      \midrule[0.6pt] 
      Qwen2.5-7B-Instruct & 49.68 & 72.12 & 85.45 & 69.08 & 54.76 & 75.04 & 68.99 & \underline{66.26} & 80.71 & 76.06 & 80.21 & 78.99 \\
      \midrule[0.6pt] 
      Liquid-7B-Qwen & 61.44 & 68.42 & 79.91 & 69.92 & 43.98 & 71.67 & 65.24 & 60.30 & 82.90 & 74.63 & 73.39 & 76.97 \\
      Genie-7B-Qwen & 64.33 & 73.73 & 80.56 & 72.87 & 48.07 & 76.70 & 66.08 & 63.62 & \underline{83.55} & 77.32 & 74.68 & 78.52 \\
      SDR-7B-Qwen & \underline{65.20} & \underline{76.73} & \underline{82.61} & \underline{74.85} & \underline{48.20} & \underline{78.25} & \underline{69.94} & 65.46 & 81.03 & \underline{79.51} & \underline{77.96} & \underline{79.50} \\
      BDS-7B-Qwen & 60.93 & 75.72 & 79.55 & 72.07  & 45.63 & 74.95 & 68.09 & 62.89 & 79.31 & 74.62 & 71.05 & 75.00 \\
      \midrule[0.6pt]
      \textbf{K2V-7B-Qwen} & \textbf{66.81} & \textbf{79.16} & \textbf{88.30} & \textbf{78.09} & \textbf{55.20} & \textbf{79.53} & \textbf{70.69} & \textbf{68.47} & \textbf{87.16} & \textbf{81.36} & \textbf{80.46} & \textbf{83.00} \\
      \midrule[0.6pt] 
      
      \multicolumn{13}{c}{\textbf{Llama-3.2-3B-Instruct (Backbone)}} \\
      \midrule[0.6pt] 
      Llama-3.2-3B-Instruct & 41.79 & 44.41 & 71.59 & 52.60 & 30.36 & 42.13 & 56.31 & 42.93 & 55.22 & 44.27 & 68.19 & 55.90 \\
      \midrule[0.6pt] 
      Liquid-3B-Llama & 51.89 & 45.52 & 69.88 & 55.76 & 31.85 & \underline{43.64} & 59.92 & 45.13 & 63.59 & 43.34 & 67.71 & 58.21 \\
      Genie-3B-Llama & 55.71 & 42.18 & 69.39 & 55.76 & 31.79 & 41.46 & \textbf{61.39} & 44.88 & 63.16 & 44.85 & 65.86 & 57.96 \\
      SDR-3B-Llama & \underline{56.41} & 48.07 & \textbf{73.16} & \underline{59.21} & \underline{33.57} & 43.57 & 58.54 & \underline{45.22} & \underline{66.16} & \underline{46.52} & \underline{70.59} & \underline{61.09} \\
      BDS-3B-Llama & 53.75 & \underline{48.41} & 65.56 & 55.91 & 30.62 & 43.14 & 61.04 & 44.93 & 64.03 & 42.47 & 61.81 & 56.10 \\
      \midrule[0.6pt] 
      \textbf{K2V-3B-Llama} & \textbf{58.60} & \textbf{50.53} & \underline{71.86} & \textbf{60.33} & \textbf{35.50} & \textbf{44.90} & \underline{61.29} & \textbf{47.23} & \textbf{68.16} & \textbf{49.43} & \textbf{71.16} & \textbf{62.92} \\
      \midrule[0.6pt] 
      
      \multicolumn{13}{c}{\textbf{Llama-3.1-8B-Instruct (Backbone)}} \\
      \midrule[0.6pt]
      Llama-3.1-8B-Instruct & 51.10 & 55.74 & 82.37 & 63.07 & 37.61 & 54.49 & \underline{70.22} & 54.11 & 68.62 & 53.55 & \textbf{79.39} & 67.19 \\
      \midrule[0.6pt] 
      Liquid-8B-Llama & 59.84 & 53.77 & \underline{81.84} & 65.15 & 40.43 & 53.25 & 68.54 & 54.07 & 69.27 & \underline{54.01} & 77.79 & \underline{67.02} \\
      Genie-8B-Llama & 62.50 & 55.49 & 77.95 & 65.31 & 41.07 & 51.17 & 70.15 & 54.13 & 68.97 & 52.41 & 76.39 & 65.92 \\
      SDR-8B-Llama & 62.11 & 54.42 & 80.22 & 65.58 & \underline{41.19} & \underline{55.22} & 66.89 & \underline{54.43} & \underline{70.26} & 53.63 & 77.08 & 66.99  \\
      BDS-8B-Llama & \underline{62.53} & \underline{56.76} & 78.75 & \underline{66.01} & 39.96 & 51.52 & 67.73 & 53.07 & 66.38 & 50.99 & 73.83 & 63.73 \\
      \midrule[0.6pt]
      \textbf{K2V-8B-Llama} & \textbf{63.90} & \textbf{58.43} & \textbf{85.14} & \textbf{69.16} & \textbf{42.35} & \textbf{57.96} & \textbf{70.79} & \textbf{57.03} & \textbf{72.55} & \textbf{56.32} & \underline{78.50} & \textbf{69.12} \\
      \midrule[0.6pt]
    \end{tabular}}
  \label{tab:knowledge-intensive results}
\end{table*}

\noindent \textbf{Checklist Synthesis.}
We propose a two-stage synthesis pipeline, as introduced below.

First, we define a set of general criteria, formally denoted as:
\begin{equation}
\label{eq:General Criteria}
G = \{g_1, g_2, \dots, g_N\}
\end{equation}
where each $g_i$ is a universal principle that characterizes a high-quality reasoning process, independent of any specific question. We developed general criteria based on the scoring rubrics from the AP Course and Exam Description \footnote{The AP Course and Exam is a program that provides a college-level introductory course curriculum to high school students.}. See Appendix~\ref{Examples of General Criteria} for the examples of general criteria.

Second, for a given QA pair consisting of a question $x$ and a ground truth $y$, we feed both $x$, $y$ and the set of general criteria $G$ into an LLM, denoted as $S$. This LLM is prompted to instantiate the general criteria $G$ into a concrete, question-specific checklist $C^x$ (see Appendix~\ref{Prompt for Synthesizing Checklist} for the prompt). This synthesis process can be expressed as:
\begin{equation}
\label{eq:synthesis}
C^x = \{c_1, \dots, c_k\} \sim S(\cdot | x, y,G)
\end{equation}
We conduct a quality assessment of the synthesized checklist, as shown in Appendix~\ref{Quality Assessment of Checklists}.

\subsection{Answer-Gated Reward Mechanism}
\label{Answer-Gated Reward Mechanism}
In order to anchor logical consistency to factual accuracy and prevent potential reward hacking, K2V employs an answer-gated reward mechanism. The total reward $R_{\text{total}}$ is defined as:
\begin{equation}
\label{eq:total_reward}
\begin{gathered}
R_{\text{total}} = R_{\text{format}} + R_{\text{answer}} + R_{\text{reason}} \\
R_{\text{reason}} = \mathbbm{1}(\hat{y} = y) \cdot p
\end{gathered}
\end{equation}
where $\mathbbm{1}(\cdot)$ is the indicator function, which equals 1 if the predicted answer $\hat{y}$ matches the ground truth $y$, and 0 otherwise. $p$ denotes the pass rate calculated in Equation~\ref{eq:pass rate}.

\noindent \textbf{Format Reward.}
The format reward $R_{\text{format}}$ encourages the policy model to generate outputs in a structured format. The desired output template is consistent with that of DeepSeek-R1~\citep{guo2025deepseek}. A maximum score of 0.75 is awarded if the output perfectly adheres to this template.

\noindent \textbf{Answer Reward.}
The answer reward $R_{\text{answer}}$ is a binary reward, where a score of $\alpha$ is awarded if the predicted answer $\hat{y}$ equals the ground truth $y$; otherwise, the score is 0. \commentZH{In our main experiments, we set $\alpha = 6$. We also conduct a sensitivity analysis on $\alpha$, as detailed in Section~\ref{Sensitivity Analysis of Answer Reward}.}

\noindent \textbf{Reasoning Reward.}
The reasoning reward $R_{\text{reason}}$ is a dense reward signal, which is 
gated by the answer's correctness. 
If the predicted answer $\hat{y}$ matches the ground truth $y$, the model receives a reasoning reward equal to the pass rate $p$. Otherwise, the reasoning reward is 0, regardless of how logical the reasoning procss appear. We conduct an ablation experiment on the answer-gated mechanism, as detailed in Section~\ref{Ablation Studies}.

\begin{table*}[t]
  \caption{Performance on general benchmarks. K2V-based models apply reinforcement learning directly to the Qwen2.5 base backbones without SFT on any general or mathematical data.}
  \centering
    \small
    \begin{tabular*}{\textwidth}{l@{\extracolsep{\fill}}ccccc}
      \toprule[0.6pt] 
      \textbf{Model} & \textbf{BBH} & \textbf{MATH-500} & \textbf{GSM8K} & \textbf{AIME2024} & \textbf{GPQA-Diamond} \\
      \midrule[0.6pt] 
       Qwen2.5-3B-Instruct & 37.21 & \textbf{66.40} & \textbf{85.05} & \phantom{0}5.00 & 31.94 \\
       \textbf{K2V-3B-Qwen} & \textbf{42.64} & 65.20 & 84.17 & \textbf{\phantom{0}5.83} & \textbf{32.45}\\
      \midrule[0.6pt] 
       Qwen2.5-7B-Instruct & 49.63 & \textbf{76.25} & \textbf{91.51} & 11.67 & 35.86 \\
       \textbf{K2V-7B-Qwen} & \textbf{55.36} & 74.60 & 88.67 & \textbf{13.33} & \textbf{36.62}\\
      \bottomrule[0.6pt]
    \end{tabular*}
  \label{tab:general results}
\end{table*}

\section{Experiments}
\label{Experiments}

\subsection{Experimental Setups}

\textbf{Domains.} We conduct experiments in three typical knowledge-intensive domains: agriculture, law, and medicine. While these domains suffer from a scarcity of high-quality data due to the high cost of expert annotation and involve open-ended text that are difficult to verify automatically, reasoning abilities are required.

\noindent \textbf{Corpora.} For agriculture, we use RiceCorpus~\citep{yang2025seedllm}, a corpus built by crawling resources from the internet, totaling approximately 37.64 MB. For law, we use DISC-Law-SFT~\citep{yue2023disc}, a dataset of unverifiable QA pairs. We directly concatenate the questions and answers, then randomly downsample 20.69 MB of the text. \commentZH{This demonstrates that K2V can convert unverifiable SFT data into verifiable data.} For medicine, we use shibing624-medical-pretrain~\footnote{https://huggingface.co/datasets/ticoAg/shibing624-medical-pretrain}, a Chinese corpus sourced from online medical encyclopedias and textbooks. We randomly downsample 23.51 MB of text from this corpus.

\noindent \textbf{LLM Backbones.} 
We adopt widely used open-source models, including Qwen2.5-3B, Qwen2.5-7B, Llama3.2-3B-Instruct, and Llama3.1-8B-Instruct, as backbones. We use them to examine the performance of K2V due to their moderate foundational capabilities in knowledge-intensive domains.

\noindent \textbf{Evaluation.} 
For agriculture, to ensure an fair evaluation, we evaluate on the objective question subset of SeedBench~\citep{ying2025seedbench} and conduct evaluations on agriculture-related subsets of CMMLU\citep{li2024cmmlu} and MMLU~\citep{hendrycks2020measuring}. For law, we evaluate on LawBench~\citep{fei2024lawbench}, as well as on law-related subsets of CMMLU and MMLU. For medicine, we evaluate on MedQA-MCMLE~\citep{jin2021disease}, as well as on medicine-related subsets of CMMLU and MMLU. The details of subset selection for all three domains are provided in the Appendix~\ref{Details of Model Evaluation}. Additionally, we conduct evaluations on general benchmarks, including BBH~\citep{suzgun2023challenging}, MATH-500~\citep{hendrycks2021measuring, lightman2023let}, GSM8K~\citep{cobbe2021training}, AIME2024~\citep{MAA_AIME_2024}, and GPQA-Diamond~\citep{rein2024gpqa}. We conduct evaluations in a zero-shot setting \commentZH{with a temperature of 0.6. All models are evaluated on each benchmark four times and the average accuracy is reported.}

\noindent \textbf{Implementation Details.} 
All experiments were conducted using the DAPO~\citep{yu2025dapo} algorithm for training. We sample 8 responses per prompt for a batch of 64 prompts. The clip threshold is set to (0.2, 0.28) and a learning rate of $1 \times 10^{-6}$. The maximum generation length and overlong buffer length are set to 4096 and 512. Both the KL divergence coefficient and the entropy regularization coefficient are set to 0.
\commentZH{Unless otherwise specified, $\mathcal{F}_\text{text}$, $S$, and the model used for KG construction are Qwen2.5-72B-Instruct, while the judge model $J$ is Qwen2.5-7B-Instruct.}
See Appendix~\ref{Impltementation Details of K2V} for more implementation details.

\noindent \textbf{Baselines.}
Following the discussion in Section~\ref{sec:related}, we compare K2V against the following baseline methods.
(1) Liquid~\citep{lee2023liquid} directly extracts questions with multiple verifiable candidate answers from corpora.
(2) Genie~\citep{yehudai2024genie} synthesizes QA pairs for standard SFT. We perform necessary modifications on its prompt to enhance the verifiability of the synthesized data.
(3) SDR~\citep{guo2025synthetic} uses a task-definition approach to synthesize verifiable QA pairs for RLVR.
(4) BDS~\citep{dedhia2025bottom} synthesizes verifiable multiple-choice data based on a KG. See Appendix~\ref{Impltementation Details of Baselines} for details of these baseline methods. In addition, we use official LLMs (\emph{i.e.}~Qwen and Llama) as the \emph{vanilla} baselines. In contrast to baseline methods, the \emph{vanilla}  baselines are not trained on the domain corpora.
\begin{figure}[t]
    \includegraphics[width=\columnwidth]{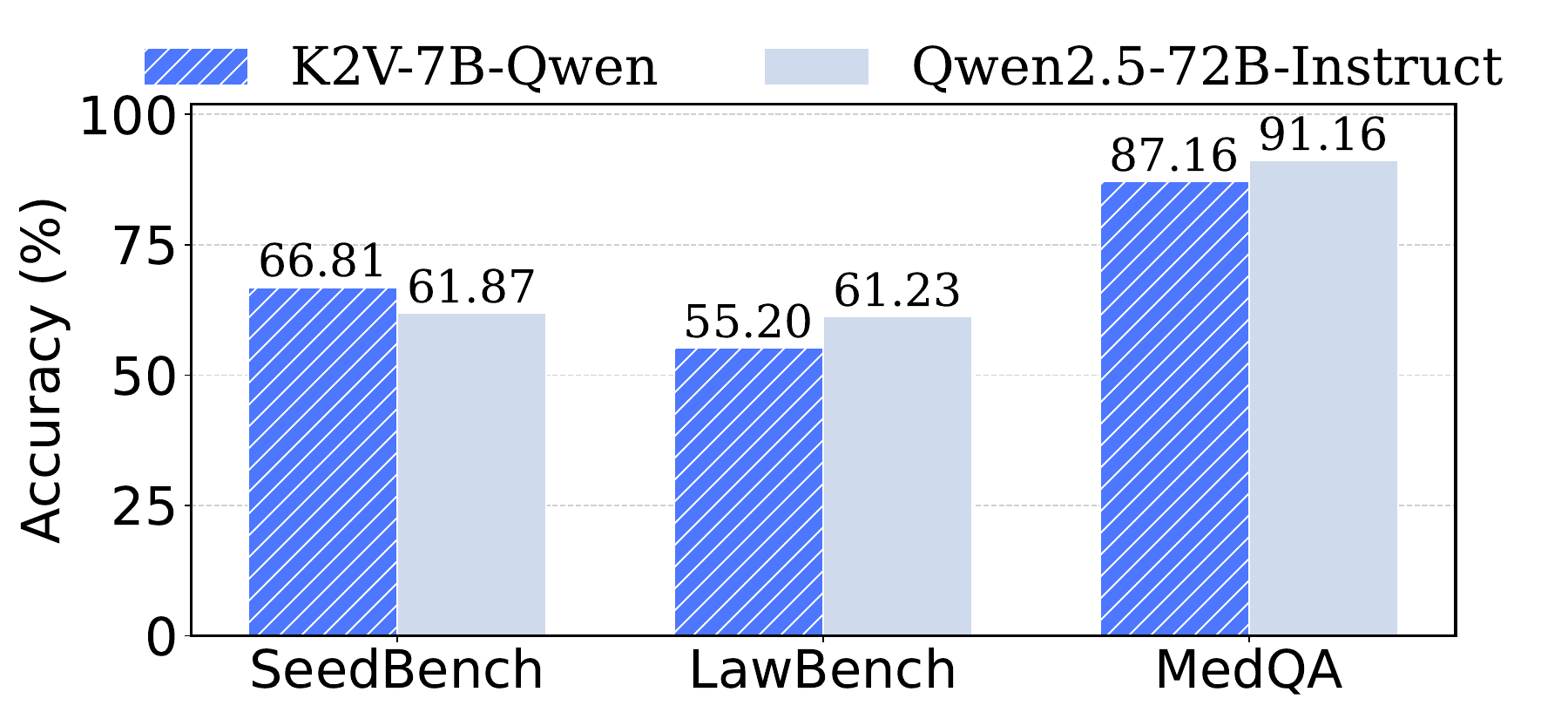}
    \caption{The accuracy of K2V-7B-Qwen and Qwen2.5-72B-Instruct in knowledge-intensive domains. A small LLM trained with K2V and domain corpora can even outperform a much larger LLM.}
    \label{fig:K2V-7B_and_Qwen25-72B}
\end{figure}
\begin{table*}[t]
  \caption{Ablation studies on components. Using K2V-3B-Qwen as the baseline, we separately ablate answer verification, reasonging verification, and answer-gated reward mechanism. SFT denotes the model fine-tuned on QA pairs synthesized by our method, using Qwen2.5-3B as the base model.}
  \centering
  \resizebox{1.0\textwidth}{!}{
    \tiny
    \setlength{\tabcolsep}{4pt}
    \begin{tabular}{l|ccc|ccccc}
      \toprule[0.6pt] 
      & \multicolumn{3}{c|}{\textbf{Knowledge-Intensive Domains}} & \multicolumn{5}{c}{\textbf{General Domains}} \\
      \multirow{-2}{*}{\textbf{Model}} &
      \textbf{SeedBench} & \textbf{LawBench} & \textbf{MedQA} & \textbf{BBH} & \textbf{MATH-500} & \textbf{GSM8K} & \textbf{AIME2024} & \textbf{GPQA-Diamond} \\
      \midrule[0.6pt] 
       \textbf{K2V-3B-Qwen} & \textbf{62.82} & \textbf{43.27} & \textbf{78.45} & \textbf{42.64} & \textbf{65.20} & \textbf{84.17} & \textbf{5.83} & \textbf{32.45} \\
       \hspace{2mm} w/o Answer Verification & 43.49 & 36.88 & 66.93 & 20.50 & 56.65 & 46.94 & 0 & 25.13 \\
       \hspace{2mm} w/o Reasoning Verification & 61.58 & 37.97 & 75.86 & 39.93 & 42.40 & 72.92 & 0.83 & 31.69 \\
       \hspace{2mm} w/o Answer-Gated & 53.35 & 35.10 & 72.18 & 35.11 & 53.85 & 58.95 & 1.67 & 28.16 \\
       \hspace{2mm} SFT & 41.38 & 30.97 & 54.12 & 30.49 & 14.80 & 20.71 & 1.67 & 25.00 \\
       \midrule[0.6pt]
    \end{tabular}}
  \label{tab:ablation_performance}
\end{table*}

\begin{figure*}[t]
    \begin{center}
        \includegraphics[width=\linewidth]{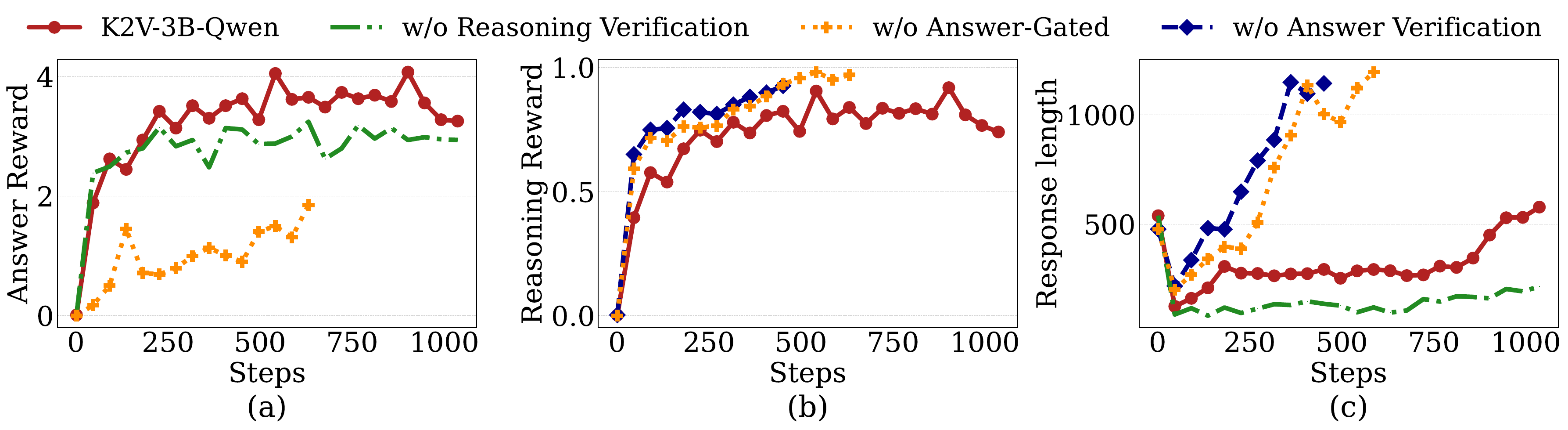}
    \end{center}
    \caption{Training dynamics of ablation studies. The \textcolor[HTML]{B22222}{\textbf{K2V-3B-Qwen}} demonstrates stable learning across all metrics. (a) and (c) show that \textcolor[HTML]{228B22}{\textbf{removing reasoning verification}} impairs the model's ability to explore correct answers. (b) and (c) show that \textcolor[HTML]{00008B}{\textbf{removing answer verification}} leads to reward hacking, where the model generates excessively long responses to maximize the reasoning reward, causing training instability. (a), (b) and (c) show that \textcolor[HTML]{FF8C00}{\textbf{removing answer-gated reward mechanism}} decouples reasoning from factual accuracy. This leads to reward hacking, where the model achieves high reasoning reward through excessive response length and learns reasoning patterns that lead to incorrect answers.}
    \label{fig:ablation_combine}
\end{figure*}

\subsection{Quantatitive Comparison with Existing RLVR Studies}
\label{Main Results}

The main experimental results on three knowledge-intensive domains are reported in Table~\ref{tab:knowledge-intensive results}, from which we present the following findings.
\textbf{Firstly}, K2V significantly enhances reasoning capabilities in knowledge-intensive domains. Across various backbone models, our method overall achieves the best average accuracy in all three domains, outperforming all baseline methods.
\textbf{Secondly}, K2V better enables a small LLM to achieve performance comparable to or even exceeding that of a much larger LLM. As shown in Figure~\ref{fig:K2V-7B_and_Qwen25-72B}, K2V-7B-Qwen exceeds Qwen2.5-72B-Instruct on SeedBench. Meanwhile, its performance on MedQA is close to that of Qwen2.5-72B-Instruct.
\textbf{Finally}, when we aim to investigate the negative effect of domain-specific training on LLM's general reasoning abilities, surprisingly, we found that K2V not only retained these skills but even improved them. As shown in Table~\ref{tab:general results}, without using any general or mathematical data, models trained with K2V do not exhibit a significant decline in general reasoning abilities compared to the \emph{vanilla} baselines, and even shows improvements on certain benchmarks. 
Specifically, K2V-3B-Qwen and K2V-7B-Qwen, which are trained directly on Qwen2.5 base backbones without prior SFT, consistently surpass the \emph{vanilla}  baselines across the BBH, AIME2024, and GPQA-Diamond. This suggests that K2V might have cross-domain advantages: the domain-specific training enables the model’s reasoning capabilities to generalize to certain general tasks.

\subsection{Ablation Studies}
\label{Ablation Studies}

\begin{table*}[htbp]
  \caption{Sensitivity analysis of the answer reward $\alpha$. Using K2V-3B-Qwen as the study object, we test $\alpha \in \{2, 4, 6, 8, 10\}$ while holding all other parameters fixed. The results across both knowledge-intensive and general domains demonstrate that K2V is robust to different magnitudes of the answer reward.}
  \centering
  \resizebox{1.0\textwidth}{!}{
    \tiny
    \setlength{\tabcolsep}{4pt}
    \begin{tabular}{l|ccc|ccccc}
      \toprule[0.6pt] 
      & \multicolumn{3}{c|}{\textbf{Knowledge-Intensive Domains}} & \multicolumn{5}{c}{\textbf{General Domains}} \\
      \multirow{-2}{*}{\textbf{$\alpha$}} &
      \textbf{SeedBench} & \textbf{LawBench} & \textbf{MedQA} & \textbf{BBH} & \textbf{MATH-500} & \textbf{GSM8K} & \textbf{AIME2024} & \textbf{GPQA-Diamond} \\
      \midrule[0.6pt] 
       $2$ & 62.45 & 42.41 & 76.30 & 42.81 & 64.95 & 81.80 & 5.83 & 29.55 \\
       $4$ & 62.82 & 43.27 & 78.78 & 42.64 & 63.60 & 82.88 & 5.83 & 32.07 \\
       $6$ & 62.82 & 43.27 & 78.45 & 42.64 & 65.20 & 84.17 & 5.83 & 32.45 \\
       $8$ & 61.98 & 42.90 & 77.29 & 42.37 & 65.90 & 81.37 & 9.17 & 30.43 \\
       $10$ & 63.20 & 41.93 & 78.69 & 41.58 & 60.95 & 81.46 & 8.33 & 31.06 \\
       \midrule[0.6pt]
    \end{tabular}}
  \label{tab:Sensitivity Analysis of Answer Reward}
\end{table*}

To investigate the individual components of K2V, we conduct comprehensive ablation studies. We compare the K2V-3B-Qwen against four variants under the same data and training parameters.

\noindent \textbf{K2V-3B-Qwen.}
This is our complete model, trained with both reasoning verification ($R_{\text{reason}}$) and answer verification ($R_{\text{answer}}$). As shown in Table~\ref{tab:ablation_performance}, this configuration consistently achieves the highest performance across both knowledge-intensive domains and general domains. The training dynamics in Figure~\ref{fig:ablation_combine} reveal a stable and effective learning process. Both the answer and reasoning reward increase steadily, while the response length follows a healthy increase pattern after an initial decrease. 

\noindent \textbf{w/o Answer Verification.} 
This variant removes the answer verification ($R_{\text{answer}}$)  and is trained solely with the reasoning verification ($R_{\text{reason}}$). As shown in Figure~\ref{fig:ablation_combine}(b), although this model initially achieves higher reasoning reward, it does so through reward hacking. 
The model discovers that longer outputs more readily meet the criteria of checklist, thus it tends to generate excessively long responses to maximize the reasoning reward.
As shown in Figure~\ref{fig:ablation_combine}(c), its response length grows uncontrollably, leading to out-of-memory (OOM) errors that halted training around 500 steps. 

\noindent \textbf{w/o Reasoning Verification.}
This variant removes the reasoning verification ($R_{\text{reason}}$) and is trained solely with the answer verification ($R_{\text{answer}}$), making it akin to a conventional RLVR. Figure~\ref{fig:ablation_combine}(a) shows that without the guidance from reasoning verification, the model achieves a lower answer reward. This demonstrates that the dense signal from $R_{\text{reason}}$ is vital for accelerating the model's learning helping it efficiently explore correct answer. Furthermore, as shown in Figure~\ref{fig:ablation_combine}(c), its response length stagnates after an initial drop, suggesting a less effective learning.

\noindent \textbf{w/o Answer-Gated.}
This variant removes the answer-gated reward mechanism, meaning the reasoning reward $R_{\text{reason}}$ is no longer gated by the final answer. Regardless of whether the predicted answer is correct, the $R_{\text{reason}}$ is set to the pass rate $p$ calculated in Equation~\ref{eq:pass rate}.
As shown in Table~\ref{tab:ablation_performance}, this model suffers from a significant performance degradation across all benchmarks. 
Although Figure~\ref{fig:ablation_combine}(b) shows the model achieves high reasoning rewards, its answer reward remains at a low level in Figure~\ref{fig:ablation_combine}(a). This indicates that the reasoning process becomes decoupled from factual accuracy, where the model learns reasoning patterns that can lead to incorrect answers. Furthermore, the rapid increase in response length in Figure~\ref{fig:ablation_combine}(c) suggests the model suffers reward hacking by generating long text to satisfy the checklist criteria. 

\noindent \textbf{SFT.}
This variant forgoes reinforcement learning entirely and instead uses the QA pairs synthesized by K2V to perform standard SFT on the Qwen2.5-3B. As detailed in Table~\ref{tab:ablation_performance}, this model suffers a significant performance collapse on all benchmarks. We hypothesize that this is because the highly structured, short-answer format data synthesized by K2V is ill-suited for SFT, which typically relies on more diverse and conversational data. SFT on verifiable QA pairs merely teaches the model a superficial mapping from questions to answers, rather than enabling it to learn deeper reasoning skills.

\subsection{Sensitivity Analysis of Answer Reward $\alpha$}
\label{Sensitivity Analysis of Answer Reward}

As shown in Equation~\ref{eq:total_reward}, the answer reward $R_{\text{answer}}$ is set to the variable $\alpha$ when the model predicts the correct final answer. 
To investigate the impact of the different answer reward magnitude, we conduct a sensitivity analysis. Specifically, we test $\alpha \in \{2, 4, 6, 8, 10\}$ on K2V-3B-Qwen while holding all other parameters fixed.

The results in Table~\ref{tab:Sensitivity Analysis of Answer Reward} demonstrate that K2V is insensitive to $\alpha$ and maintains stable performance across all configurations. This robustness is primarily attributed to DAPO’s group reward normalization, which focuses on the relative ranking of responses for a rollout rather than absolute reward magnitudes. Additionally, we believe that the dense reasoning reward signals provided by checklist-style verification effectively guide the model's exploration, reducing its reliance on answer rewards.



\begin{table*}[htbp]
  \caption{Results of smaller generator. We use three instruction-tuned versions of Qwen2.5 models (72B, 14B, and 7B) as generators to perform the entire pipeline of data synthesis. The data synthesized by each generator is then used to train Qwen2.5-3B.}
  \centering
  \resizebox{1.0\textwidth}{!}{
    \tiny
    \setlength{\tabcolsep}{4pt}
    \begin{tabular}{l|ccc|ccccc}
      \toprule[0.6pt] 
      & \multicolumn{3}{c|}{\textbf{Knowledge-Intensive Domains}} & \multicolumn{5}{c}{\textbf{General Domains}} \\
      \multirow{-2}{*}{\textbf{Generator size}} &
      \textbf{SeedBench} & \textbf{LawBench} & \textbf{MedQA} & \textbf{BBH} & \textbf{MATH-500} & \textbf{GSM8K} & \textbf{AIME2024} & \textbf{GPQA-Diamond} \\
      \midrule[0.6pt] 
       72B & 62.82 & 43.27 & 78.45 & 42.64 & 65.20 & 84.17 & 5.83 & 32.45 \\
       14B & 59.71 & 40.62 & 76.16 & 41.36 & 62.40 & 83.24 & 5.83 & 31.31 \\
       7B & 60.94 & 42.79 & 76.74 & 41.29 & 63.00 & 83.85 & 5.83 & 31.44 \\
       \midrule[0.6pt]
    \end{tabular}}
  \label{tab:Results of Smaller Generator}
\end{table*}

\subsection{Results of Smaller Generator}
Our approach relies on the in-context understanding ability of LLMs for data synthesis. Specifically, we employ Qwen2.5-72B-Instruct as a generator to convert the masked quintuplets into fill-blank style QA pairs ($\mathcal{F}_\text{text}$), instantiate the general criteria ($S$), and construct the KG. To investigate whether smaller models can serve as generators for data synthesis, We replace the 72B generator with Qwen2.5-14B-Instruct and Qwen2.5-7B-Instruct to perform the entire pipeline of data synthesis. 
We then use Qwen2.5-3B as the backbone to conduct experiments on the data synthesized by these different generators.
For a fair comparison, all training parameters are kept identical to those in the main experiment, and the judge model $J$ remains Qwen2.5-7B-Instruct. 

As shown in Table~\ref{tab:Results of Smaller Generator}, using a 7B or 14B model as the generator results in only a slight performance drop compared to the 72B model. This demonstrates that K2V is robust to the size of generator and can synthesize high-quality data without strictly relying on large generators.

\subsection{Computational Cost of Data Synthesis}
Our method relies on LLMs throughout the entire pipeline of data synthesis, including KG construction, QA pairs synthesis, and checklist synthesis. To ensure the cost is manageable, we report the computational cost (GPU hours on one H100) of the LLM-related stages when using Qwen2.5-72B-Instruct as the generator. As shown in Table~\ref{tab:Computational Cost of Data Synthesis}, even with a relatively large model as the generator, the computational cost of K2V for data synthesis remains within a generally acceptable range.

\begin{table}[t]
\centering
\caption{Computational cost of data synthesis. We report the computational cost (GPU hours on one H100) when the generator is Qwen2.5-72B-Instruct.}
\label{tab:Computational Cost of Data Synthesis}
\begin{tabular}{lc}
\midrule[0.6pt]
\textbf{Stage} & \textbf{GPU hours} \\ 
\midrule[0.6pt] 
KG construction & 201.25 \\
QA pairs synthesis & 27.00 \\
Checklist synthesis & 53.34 \\
Overall & 281.59 \\                             
\midrule[0.6pt] 
\end{tabular}
\end{table}

\section{Conclusion}
In this paper, we present K2V, a novel framework that explores RLVR in knowledge-intensive domains. 
First, we develop fill-blank style verification to automatically synthesize verifiable QA pairs. 
Moreover, we introduce a checklist-style verification method that not only verifies the model's reasoning process but also provides a dense reward signal. 
Finally, we propose an answer-gated reward mechanism to integrate answer reward and reasoning reward, which ensures the factual correctness and prevents potential reward hacking.
Comprehensive experimental results show that K2V improves reasoning capabilities in knowledge-intensive domains without significantly compromising general capabilities. 
In the future, we aim to study the multimodal verifiable data synthesis, which plays a critical role in knowledge-intensive reasoning.

\section*{Limitations}
Despite its effectiveness, this work has a few limitations. First, due to computational resource constraints, our experiments were primarily focused on small to medium-sized models, such as the 3B and 8B versions of Qwen and Llama; therefore, the scalability of K2V on ultra-large-scale models (e.g., 70B parameters) remains to be further explored. Second, the current framework is specifically designed to address the lack of verifiable data in knowledge-intensive domains like agriculture, law, and medicine. Consequently, we did not extend our synthesis and verification methods to the fields of mathematics and coding, which already have relatively mature verification mechanisms. Future research could focus on integrating these diverse domains into a more unified RLVR framework and evaluating performance across a broader range of model scales.

\section*{Acknowledgments}
This work was supported by the New Generation Artificial Intelligence-National Science and Technology Major Project (2025ZD0121801). 


\bibliography{custom}

\clearpage
\appendix

\section{Details of Model Evaluation}
\label{Details of Model Evaluation}

As shown in Section~\ref{Experiments}, to ensure an objective and fair evaluation, we evaluate the model on the objective subset of SeedBench. Specifically, we used six subsets including 1-1, 1-2, 1-3, 3-1, 3-2, and 3-3. These subsets have unique ground-truth answers, ensuring the fairness of the evaluation.

Furthermore, to conduct a more comprehensive evaluation in the domains of agriculture, law, and medicine, we also select corresponding subsets from CMMLU and MMLU. The specific subsets are as follows:
\begin{itemize}
  \item CMMLU in the agricultural: virology, high school biology, food science, agronomy.
  \item CMMLU in the law: legal \& moral basis, professional law, jurisprudence, 
  international law, college law.
  \item CMMLU in the medicine: anatomy, high school biology, medical statistics, virology, college medicine, Clinical knowledge, professional medicine, traditional, chinese medicine.
  
  \item MMLU in the agricultural: high school biology, college biology.
  \item MMLU in the law: international law, jurisprudence, professional law.
  \item MMLU in the medicine: anatomy,
  clinical knowledge, high school biology, 
  college medicine, medical genetics, 
  professional medicine, college biology.
\end{itemize}

We use OpenCompass ~\citep{2023opencompass} as the evaluation framework. All evaluations were conducted in a zero-shot setting. We perform four inference runs for each benchmark and reported the Avg@4 results. We employ a temperature of 0.6.

\section{Impltementation Details of K2V}
\label{Impltementation Details of K2V}

In this section, we describe the implementation details of our work, including data synthesis and model training.

\subsection{Data Synthesis}
K2V first constructs a knowledge graph (KG) using the Graphgen~\citep{chen2025graphgen} framework (parameters are listed in Table~\ref{tab:graphgen}). Graphgen is a framework designed for synthesizing SFT data that utilizes LLMs for Entity Recognition (ER) and Named Entity Recognition (NER); we modified its code to suit the requirements of K2V. Next, K2V samples quintuplets from the knowledge graph and randomly masks an entity. An LLM then converts these masked quintuplets into verifiable question-answer (QA) pairs in a fill--blank format. Finally, K2V generates a checklist for each QA pair using an LLM. All steps involving LLMs—from KG construction to QA pairs synthesis and checklist generation—employ Qwen2.5-72B-Instruct.

\subsection{Model Training}
\label{Model Training}

All experiments were implemented using the DAPO algorithm \citep{yu2025dapo} based on the verl framework~\citep{sheng2025hybridflow}. For each batch of 64 prompts, we sampled 8 responses per prompt. The clipping threshold and learning rate were set to (0.2, 0.28) and $1 \times 10^{-6}$. The number of training epochs is set to 2. The maximum generation and overlong buffer lengths were fixed at 4,096 and 512, respectively. 
To ensure high-quality training data, we employ the dynamic sampling strategy with group filtering. Specifically, we generate candidates from a prompt batch that is three times larger than the training batch (i.e., a generation batch size of 192) and filter them based on sequence rewards. For the optimization objective, both the KL divergence coefficient and the entropy regularization coefficient are set to 0. To accelerate the rollout phase, we utilize the vLLM engine. Furthermore, to optimize GPU memory usage, we enable Fully Sharded Data Parallel (FSDP) with parameter and optimizer offloading, as well as gradient checkpointing. The judge model $J$ is Qwen2.5-7B-Instruct. All experiments were conducted on 8 Nvidia H100 GPUs. A summary of the key hyperparameters can be found in Table~\ref{tab:training_hyperparams}.

\begin{table}[t]
    \centering
    \caption{Key hyperparameters for model training in verl.}
    \label{tab:training_hyperparams}
    \small
    \begin{tabular}{ll}
        \toprule
        \textbf{Hyperparameter} & \textbf{Value} \\
        \midrule
        \multicolumn{2}{l}{\textit{Optimization \& Algorithm}} \\
        \hspace{3mm} Algorithm & DAPO \\
        \hspace{3mm} Learning rate & $1 \times 10^{-6}$ \\
        \hspace{3mm} Clip ratio range (asymmetric) & $[0.2, 0.28]$ \\
        \hspace{3mm} KL coefficient & $0.0$ \\
        \hspace{3mm} Entropy coefficient & $0.0$ \\
        \hspace{3mm} Overlong buffer length & $512$ \\
        \hspace{3mm} Overlong penalty factor & $1.0$ \\
        \midrule
        \multicolumn{2}{l}{\textit{Data \& Generation}} \\
        \hspace{3mm} Global batch size & $64$ \\
        \hspace{3mm} Mini-batch size & $32$ \\
        \hspace{3mm} Micro-batch size (per GPU) & $16$ \\
        \hspace{3mm} Generation batch size & $192$ \\
        \hspace{3mm} Rollout samples per prompt ($N$) & $8$ \\
        \hspace{3mm} Max prompt length & $1024$ \\
        \hspace{3mm} Max response length & $4096$ \\
        \midrule
        \multicolumn{2}{l}{\textit{System \& Infrastructure}} \\
        \hspace{3mm} Inference engine & vLLM \\
        \hspace{3mm} FSDP offload (Param \& Optim) & True \\
        \hspace{3mm} Gradient checkpointing & True \\
        \hspace{3mm} GPU memory utilization & $0.8$ \\
        \bottomrule
    \end{tabular}
\end{table}

\begin{table*}[t] 
    \centering 
    \caption{The hyperparameters of GraphGen, which is used to construct the KG.}
    \label{tab:graphgen}
    \begin{tabular}{l l p{7cm}}
        \toprule
        \textbf{Parameter} & \textbf{Our Value} & \textbf{Description} \\
        \midrule
        qa\_form & aggregated & Type of QA form desired. \\
        expand\_method & max\_width & Method for controlling graph expansion. \\
        \multirow{2}{*}{bidirectional} & \multirow{2}{*}{True} & Expanding the graph in both directions (True) or one direction (False). \\
        max\_extra\_edges & 2 & Maximum number of edges to expand. \\
        max\_tokens & 256 & Maximum number of tokens. \\
        max\_depth & 1 & Maximum depth for traversal in each direction. \\
        edge\_sampling & max\_loss & Strategy for edge selection at the same layer. \\
        isolated\_node\_strategy & ignore & Handling strategy for isolated nodes. \\
        \bottomrule
    \end{tabular}
\end{table*}

\section{Impltementation Details of Baselines}
\label{Impltementation Details of Baselines}

\textbf{Liquid.}
We reproduced the Liquid~\citep{lee2023liquid} method as our baseline, which consists of four stages: answer extraction, question generation, iterative filtering, and answer expansion. For entity recognition, we used spaCy~\citep{Honnibal_spaCy_Industrial-strength_Natural_2020} and BERN2~\citep{sung2022bern2} to identify general-domain and biomedical entities respectively, employing the same corpus as our K2V method. While default models were applied in all model-dependent stages, we replaced them with structurally similar Chinese-adapted models for Chinese corpus processing, as the original models do not support Chinese. All training configurations align with those used in K2V.

\noindent \textbf{Genie.}
We re-implemented Genie~\citep{yehudai2024genie} as a baseline, following its three-stage pipeline of content preparation, generation, and filtering. To ensure a fair comparison with our K2V method, we employ Qwen2.5-72B-Instruct for the generation stage. We slightly modify the original prompts to improve the verifiability of the generated data, making it better suited for RLVR. Since the official source code is not publicly available, we independently implemented the filtering mechanism based on the methodology described in the paper, including checks for format, faithfulness, and quality. All training settings follow those of K2V.

\noindent \textbf{Synthetic Data RL.}
We adopted the data generation pipeline from the Synthetic Data RL (SDR)~\citep{guo2025synthetic} framework as a comparative baseline. Utilizing the official implementation , we employed Qwen2.5-72B-Instruct as the instructor model for data synthesis and rewriting while Qwen2.5-7B-Base served as the reference model to assess sample difficulty. We customized the task description and format instructions to suit the agriculture, legal, and medical domains and grounded the synthesis process by retrieving relevant context from our proprietary raw corpora. We strictly adhered to the original protocol by performing initial generation, difficulty-adaptive rewriting, and consistency-based filtering to isolate high-potential samples. All training settings follow those of K2V.
\\
\noindent \textbf{BDS.}
We re-implemented Bottom-up Domain-specific Superintelligence (BDS)~\citep{dedhia2025bottom} as a baseline, which consists of three stages:
(1) Content Preparation: We construct a knowledge graph and systematically extract 4-node simple paths (3-hop relations) as logical chains for question generation. To ensure computational feasibility, we limit the maximum number of paths to 20,000 and restrict source nodes to those with outgoing edges, considering the first 1,000 such nodes.
(2) Generation: We employ Qwen2.5-72B-Instruct for the generation stage, consistent with our K2V method settings. For each extracted path, we construct a prompt that requests a question in multiple-choice format (4 options). 
(3) Filtering: Following the description in the BDS paper, We re-design a rule-based post-processing function to enforce strict quality control. This parser validates the output structure, extracts question stems, options, and correct answers, and automatically discards malformed generations.
All training configurations align with those used in K2V.

\section{Results Across Diverse Backbones}
\label{Results Across Diverse Backbone}

\begin{table*}[th]
  \caption{Results across diverse models. K2V-3B-Qwen and K2V-3B-Qwen-Ins use Qwen2.5-3B and Qwen2.5-3B-Instruct as their backbone models, respectively. Similarly, K2V-3B-Llama and K2V-3B-Llama-Base are built upon the Llama-3.2-3B-Instruct and Llama-3.2-3B, respectively. The performance collapse of K2V-3B-Llama-Base is not attributable to K2V, but rather to the RL-unfriendly characteristics of Llama-3.2-3B. This base model has significant distributional gap between pretraining data and reasoning tasks, which causes the model to exhibit anomalous behaviors, such as prematurely generating answers or falling into infinite loops. In fact, this performance collapse is not unique to K2V; several other baseline methods encounter similar issues on the Llama-3.2-3B.}
  \centering
  \resizebox{\textwidth}{!}{
    \large
    \begin{tabular}{l|cccc|cccc|cccc}
      \toprule[0.6pt] 
      & \multicolumn{4}{c|}{Agriculture} & 
      \multicolumn{4}{c|}{Law} & \multicolumn{4}{c}{Medicine}  \\
      \multirow{-2}{*}{\textbf{Model}} &  
      \textbf{SeedBench} & \textbf{CMMLU} & \textbf{MMLU} & \textbf{Avg} &
      \textbf{LawBench} & \textbf{CMMLU} & \textbf{MMLU} & \textbf{Avg} & \textbf{MedQA} & \textbf{CMMLU} & \textbf{MMLU} & \textbf{Avg}
      \\
      \midrule[0.6pt] 
       \multicolumn{13}{c}{\textbf{Qwen models}} \\
      \midrule[0.6pt] 
      Qwen2.5-3B-Instruct & 45.67 & 61.23 & 73.94 & 60.28 & 42.39 & 62.82 & 63.10 & 56.10 & 73.01 & 61.22 & 69.67 & 67.97 \\
      K2V-3B-Qwen & \textbf{62.82} & 66.82 & 75.40 & 68.34 & \textbf{43.27} & \textbf{71.55} & 62.01 & 58.94 & \textbf{78.45} & \textbf{67.53} & 70.76 & 72.24 \\ 
      \textbf{K2V-3B-Qwen-Ins} & 61.38 & \textbf{67.82} & \textbf{76.90} & \textbf{68.70} & 45.57 & 65.28 & \textbf{63.59} & 58.14 & 76.39 & 65.94 & \textbf{71.51} & 71.28 \\

      \midrule[0.6pt] 
      \multicolumn{13}{c}{\textbf{Llama models}} \\
      \midrule[0.6pt] 
      Llama-3.2-3B-Instruct & 41.79 & 44.41 & 71.59 & 52.60 & 30.36 & 42.13 & 56.31 & 42.93 & 55.22 & 44.27 & 68.19 & 55.90 \\ 
      K2V-3B-Llama & \textbf{58.60} & \textbf{50.53} & \textbf{71.86} & \textbf{60.33} & \textbf{35.50} & \textbf{44.90} & \textbf{61.29} & \textbf{47.23} & \textbf{68.16} & \textbf{49.43} & \textbf{71.16} & \textbf{62.92} \\
      \textbf{K2V-3B-Llama-Base} & 28.83 & 19.91 & 29.56 & 26.10 & 10.66 & 13.22 & 21.66 & 15.18 & 25.29 & 16.44 & 26.35 & 22.69 \\
      \midrule[0.6pt] 
    \end{tabular}}
  \label{tab:knowledge-intensive results across diverse backbone}
\end{table*}

In Section~\ref{Main Results}, we conduct experiments in three knowledge-intensive domains using the Qwen base models (Qwen2.5-3B and Qwen2.5-7B) and the Llama instruction-tuned models (Llama-3.2-3B-Instruct and Llama-3.1-8B-Instruct). 

We also conduct experiments using the Qwen instruction-tuned model (Qwen2.5-3B-Instruct) and the Llama base model (Llama-3.2-3B). As shown in Table~\ref{tab:knowledge-intensive results across diverse backbone}, K2V-3B-Llama-Base exhibits a severe performance collapse. It is important to emphasize that this failure is not attributable to flaws within the K2V framework itself, but is a direct consequence of the RL-unfriendly nature of the Llama-3.2-3B. As analyzed by OctoThinker~\citep{wang2025octothinker}, Llama base models demonstrate instability during RL due to a significant gap between their pretraining data distribution and the requirements of reasoning tasks. This disparity causes the model to produce anomalous behaviors, such as premature answer generation or falling into infinite loops. The robustness of K2V has been demonstrated by the success of Qwen2.5-3B and Llama-3.2-3B-Instruct. The primary bottleneck lies in the Llama base model’s lack of necessary cold-start.

This performance gap suggests that an RL-friendly base model is essential for scaling reasoning capabilities. Although K2V provides high-quality verifiable data and fine-grained rewards, these advantages cannot compensate for a base model's inability to maintain coherent reasoning. The improvement of Llama-3.2-3B-Instruct and Qwen2.5-3B demonstrates K2V’s strong generalization across different architectures. The failure of the Llama-3.2-3B base model likely stems from its inability to support direct reinforcement learning without a cold-start.

\section{Analysis of Data Leakage}
\label{Data Leakage Analysis}

To ensure that the performance gains of K2V are not caused by memorizing test data, We checked for overlaps between K2V-synthesized questions and the SeedBench, LawBench, and MedQA benchmarks. We calculate $n$-gram similarity~\citep{alneyadi2016survey} using the Qwen2.5 tokenizer, testing $n$ values from 22 to 30. A test sample is considered leaked if any of its $n$-token sequences appear in the training set. Table~\ref{tab:leakage_results} shows the number of leaked samples detected, which indicate a low risk of data leakage.

\section{Quality Assessment of Checklists}
\label{Quality Assessment of Checklists}

To assess the quality of checklist, we employ Qwen3-235B-A22B-Instruct-2507 as the judge model to assess the checklists. Specifically, we randomly sample 10000 instances from the synthesized checklist across the agriculture, legal, and medical domains. The judge model evaluates each checklist on a scale of 1 to 5 based on three specific dimensions:
\begin{itemize}[itemsep=2pt, topsep=2pt, parsep=0pt]
    \item Relevance: Are the items in the checklist directly relevant to the specific question and the ground truth ? Do they check for information that actually matters for this problem ?
    \item Verifiability: Are the criteria objective and verifiable? Can a third-party evaluator easily determine ``yes'' or ``no'' without ambiguity?
    \item Necessity: Does the checklist cover the necessary steps or facts required to reach the correct conclusion? Are there missing critical steps or redundant unnecessary steps?
\end{itemize}

The assessment results in Table~\ref{tab:checklist_quality} demonstrate that the K2V is able to produce high-quality checklists across all dimensions. Specifically, the high necessity score (4.60) indicates that the checklists effectively cover the essential logical chains and knowledge points required for accurate reasoning, while the strong scores in relevance (4.37) and verifiability (4.29) confirm that the criteria are both closely aligned with the questions and sufficiently objective to provide stable reward signals. The results confirm that K2V can automatically synthesize checklists that provide a robust foundation for verifying the LLM's reasoning process.

\begin{table}[t]
\centering
\caption{Data leakage detection results across different $n$-gram lengths. The results indicate a low risk of data leakage}
\label{tab:leakage_results}
\begin{tabular}{lcccc}
\midrule[0.6pt] 
\multirow{2}{*}{\vspace{-0.6em}\textbf{Benchmark}} & \multirow{2}{*}{\vspace{-0.6em}\textbf{Total Samples}} & \multicolumn{3}{c}{\textbf{$n$}} \\ \cmidrule[0.6pt](lr){3-5} & & \textbf{$22$} & \textbf{$26$} & \textbf{$30$} \\ \midrule[0.6pt] 
SeedBench   & 1,975                  & 5               & 0               & 0               \\
LawBench    & 9,500                  & 30              & 20              & 1               \\
MedQA       & 3,426                  & 0               & 0               & 0               \\ \midrule[0.6pt] 
\end{tabular}
\end{table}

\begin{table}[t]
\centering
\caption{Average quality scores of synthesized checklists. We randomly sample 10000 instances from the synthetic checklists and assess their quality using Qwen3-235B-A22B-Instruct-2507 across three dimensions: relevance, verifiability, and necessity. Score denotes the average value across these 10000 instances, with a range of [1, 5].}
\label{tab:checklist_quality}
\begin{tabular}{lc}
\midrule[0.6pt]
\textbf{Evaluation Dimension} & \textbf{Score} \\ 
\midrule[0.6pt] 
Relevance                     & 4.37                         \\
Verifiability                 & 4.29                         \\
Necessity                     & 4.60                         \\
\midrule[0.6pt]
\end{tabular}
\end{table}

\section{Quality Assessment of Knowledge Graph}
\label{Quality Assessment of Knowledge Graph}
To comprehensively validate the quality of the KG, we conduct both automated and manual evaluation of the constructed KG.

\subsection{Automated Evaluation}
Following the previous work of quality evaluation on KG~\citep{lukarma, Pan_2024}, we conduct evaluation on three dimensions: extraction accuracy, semantic consistency, and structural robustness.

\subsubsection{Extraction Accuracy}
Extraction accuracy measures the quality of named entity recognition (NER) and relation extraction (RE). 
Based on the LLM-as-judge method~\citep{ZhengC00WZL0LXZ23}, we employ Qwen2.5-72B-Instruct to compute the three metrics: (1) Accuracy: Measures the proportion of correct entities among all extracted entities. (2) Completeness: Measures the proportion of ground-truth entities successfully extracted from each text chunk. (3) Precision: Measures the exactness of entity names and descriptions.

As shown in Table~\ref{tab:Extraction Accuracy}, our KG construction pipeline demonstrates high extraction quality. Due to time and resource constraints, we randomly sampled 1000 text chunks to evaluate the extraction accuracy.

\subsubsection{Semantic Consistency}
Semantic consistency measures the degree of semantic conflict among multi-source entities, which refer to entities appearing in multiple text chunks simultaneously. We calculate two types of conflicts: (1) Entity Type Consistency: An entity is considered to be in conflict if its type differs across different chunks. (2) Description Consistency: An entity is considered to be in conflict if its description across different chunks exhibit discrepancies. We employ an LLM-as-Judge approach to detect such conflicts. Evaluation results show that our method achieves a low conflict rate(1.52\%).

\subsubsection{Structural Robustness}
Structural robustness measures the structural integrity of KG. We compute two metrics that represent characteristics of the KG: (1) Noise Ratio: Measures the proportion of isolated nodes (entities without any relationships). Lower values indicate better quality. (2) Connectivity (LCC-Ratio): Measures the proportion of nodes contained within the largest connected component. Higher values indicate better quality. Evaluation results show that our method exhibits low noise ratio (Noise Ratio=0.088) and robust connectivity (LCC-Ratio=0.859).

\begin{table}[t]
\centering
\caption{The result of the extraction accuracy of constructed KG.}
\label{tab:Extraction Accuracy}
\begin{tabular}{lcc}
\midrule[0.6pt]
\textbf{Metric} & \textbf{NER} & \textbf{RE} \\ 
\midrule[0.6pt] 
Accuracy & 0.726 & 0.810 \\
Completeness & 0.669 & 0.725 \\
Precision & 0.721 & 0.782 \\                         
\midrule[0.6pt] 
\end{tabular}
\end{table}

\subsection{Manual Evaluation}
To further validate the factual accuracy of constructed KG, we conduct a manual evaluation. Specifically, we first randomly sample 200 entities and 200 relations from the KG, where each entity or relation corresponds to a raw text chunk. Secondly, two PhD students manually verify whether the description of sampled entities and relations are consistent with the factual knowledge contained in the corresponding text chunks. Finally, we calculate the consistency rate $CR$ for the sampled entities and relations using the following formula:
\begin{equation}
\label{eq:Manual Evaluation of KG}
CR = \frac{M}{N}
\end{equation}
Where $N$ is the total number of entities or relations. $M$ is the number of entities or relations that are consistent with the raw text chunk.

Evaluation results show that our method yields a consistency rate of 97.00\% for entities and 95.50\% for relations. These high consistency rates ensure the factual accuracy of constructed KG.

\section{Qualitative Comparison with Existing RLVR Methods}
\label{Case Study}

In this section, we present the QA pairs synthesized by K2V, Liquid, Genie, SDR, and BDS, and provide an analysis of these cases.

\subsection{Synthesized Data of K2V}
\begin{figure*}[t]
    \begin{center}
        \includegraphics[width=\linewidth]{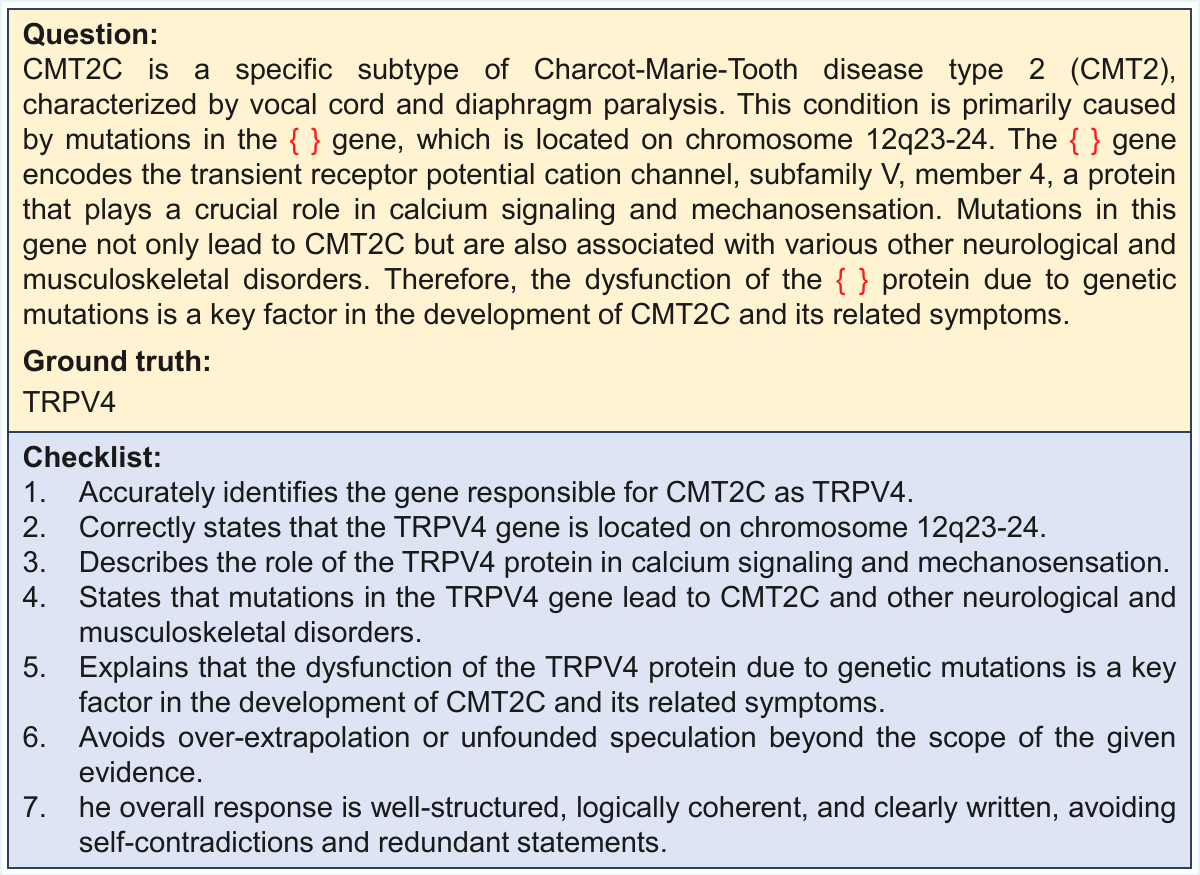}
    \end{center}
    \caption{The first case of K2V.}
    \label{fig:human_result1}
\end{figure*}

The first case is shown in Figure~\ref{fig:human_result1}. 
The question of this case provides a rich clinical and molecular context that necessitates complex medical reasoning. Rather than a simple factual query, the question integrates specific symptoms of CMT2C, such as vocal cord and diaphragm paralysis, with precise genomic data like chromosome 12q23-24 and the protein's role in calcium signaling.
The checklist of this case effectively validates whether the model understands the underlying pathophysiology of CMT2C by requiring not only the correct identification of the TRPV4 gene but also its specific chromosomal location (12q23-24) and its biological role in calcium signaling and mechanosensation. This checklist prevents the model from relying on simple keyword matching, as it must explain the causal link between genetic mutations and protein dysfunction to satisfy the criteria. 

\begin{figure*}[t]
    \begin{center}
        \includegraphics[width=\linewidth]{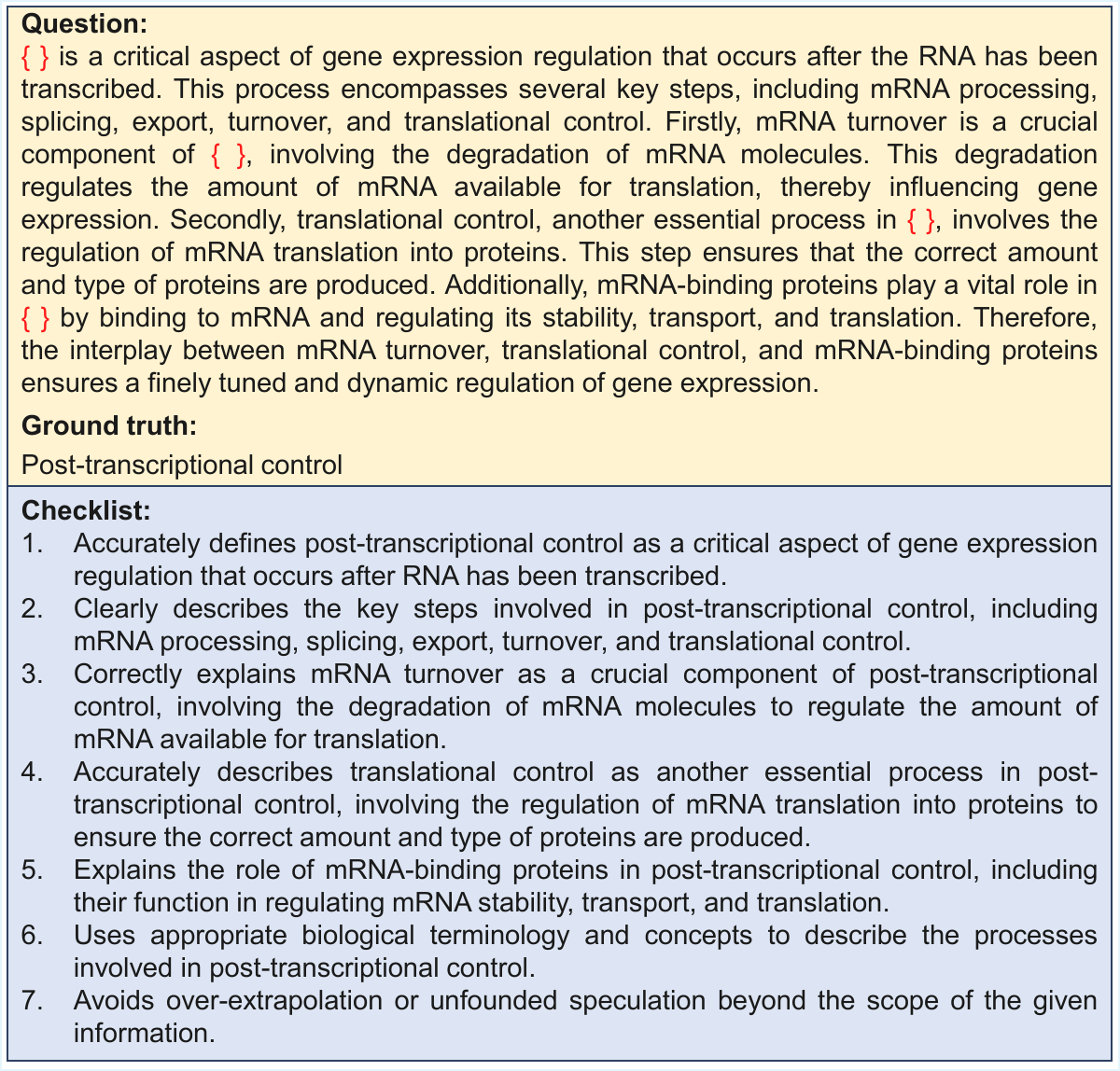}
    \end{center}
    \caption{The second case of K2V.}
    \label{fig:human_result2}
\end{figure*}

The second case is shown in Figure~\ref{fig:human_result2}. 
The question of this case is more than a simple definition lookup. By describing the interactions between multiple subprocesses, such as mRNA turnover, translational control, and mRNA-binding proteins, it constructs a complex knowledge context. The model must understand both the temporal dimension (occurring after transcription) and the functional dimension (regulation of stability and translation) to successfully infer and reconstruct the core concept of post-transcriptional control.
The checklist of this case ensures reasoning completeness by requiring not only a standard definition but also a clear description of essential steps such as mRNA processing, splicing, and export. In terms of technical depth, the checklist specifically mandates explanations for mRNA turnover and translational control, effectively verifying whether the model understands how these mechanisms co-regulate final protein levels. 

\begin{figure*}[t]
    \begin{center}
        \includegraphics[width=\linewidth]{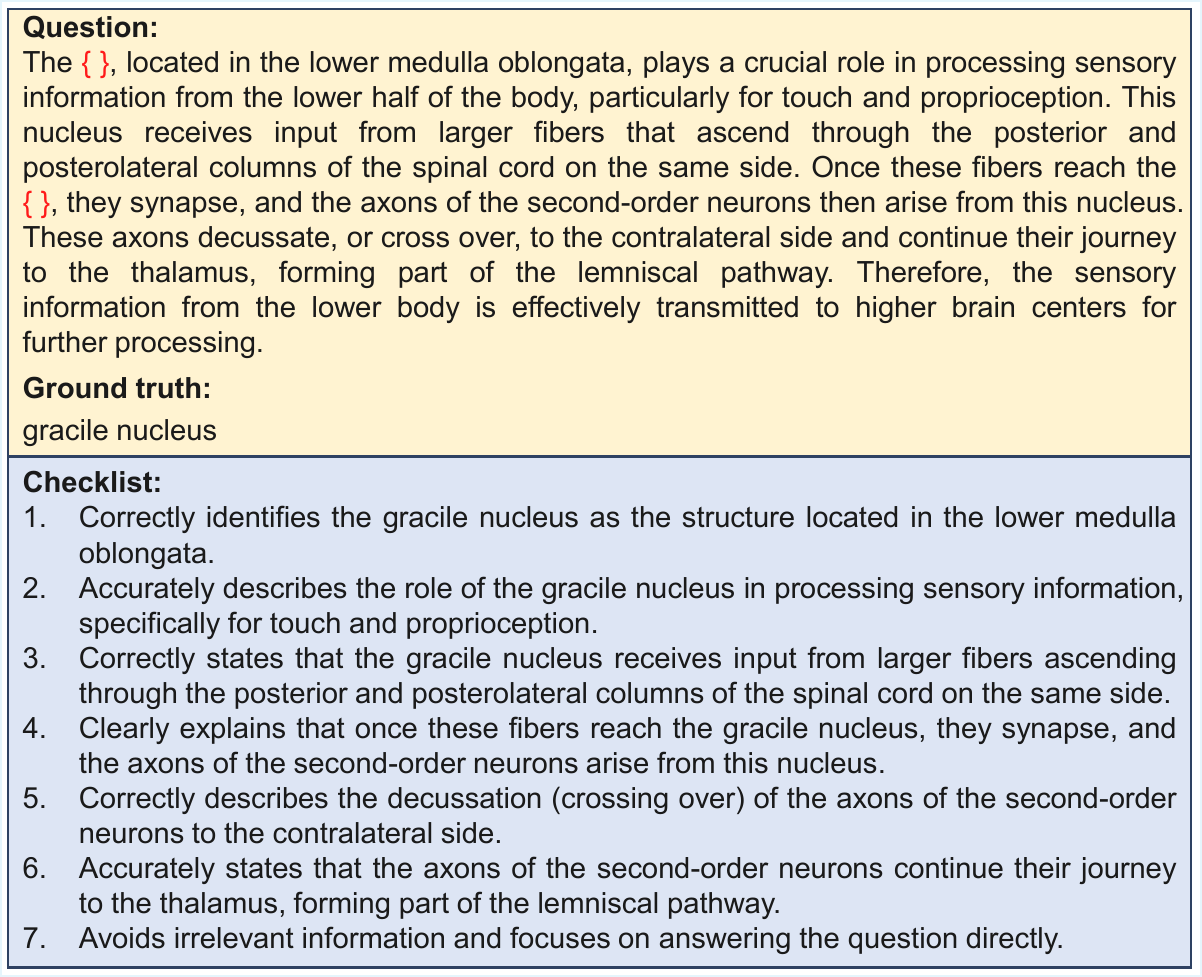}
    \end{center}
    \caption{The third case of K2V.}
    \label{fig:human_result3}
\end{figure*}

The third case is shown in Figure~\ref{fig:human_result3}. 
The question of this case is not a simple definition of terms. Instead, it establishes a dense contextual framework by detailing the anatomy of the lower medulla, the sensory pathways of the lower body, and the synapsing and decussation of second-order neurons. This narrative requires the model to accurately distinguish the "gracile nucleus" (for the lower body) from the "cuneate nucleus," thereby validating its deep understanding of hierarchical neural pathways.
The checklist of this case does not merely repeat the answer; instead, it decomposes the complex physiological process of the Dorsal Column-Medial Lemniscus (DCML) pathway into key logical nodes. First, it accurately captures anatomical specificity by verifying whether the model correctly associates the gracile nucleus with lower-body sensation, as opposed to the cuneate nucleus of the upper body. Second, it covers the dynamics of sensory transmission in detail, from the ipsilateral ascent of first-order neurons in the spinal cord to the synapse in the medulla and the subsequent decussation to the contralateral thalamus. 

\subsection{Synthesized Data of Liquid}

\begin{figure*}[t]
    \begin{center}
        \includegraphics[width=\linewidth]{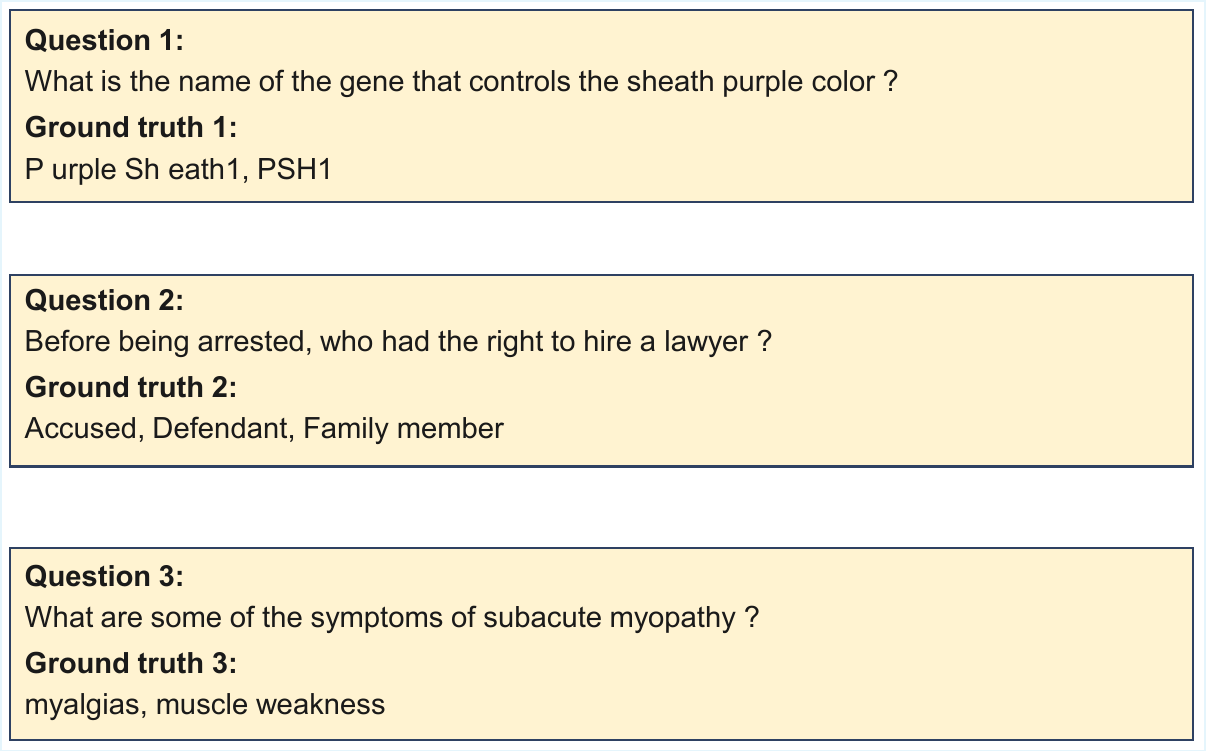}
    \end{center}
    \caption{Three cases of Liquid. The data synthesized by Liquid contains multiple candidate answers, which are separated by commas in the figure.}
    \label{fig:case_study_liquid}
\end{figure*}

As shown in Figure~\ref{fig:case_study_liquid}, the data synthesized by Liquid primarily consists of simple, direct fact-retrieval questions that lack the rich contextual constraints necessary to stimulate complex reasoning chains. While these QA pairs cover domain-specific terms, their brevity limits the model's need to perform multi-step logic or integrate diverse knowledge points, which are essential for developing sophisticated reasoning capabilities. Furthermore, the ground truths often provide a list of synonymous or categorical terms. In knowledge-intensive fields like law or medicine, such relatively shallow tasks may lead the model to prioritize surface-level memorization over logical deduction.

\subsection{Synthesized Data of Genie}
\begin{figure*}[t]
    \begin{center}
        \includegraphics[width=\linewidth]{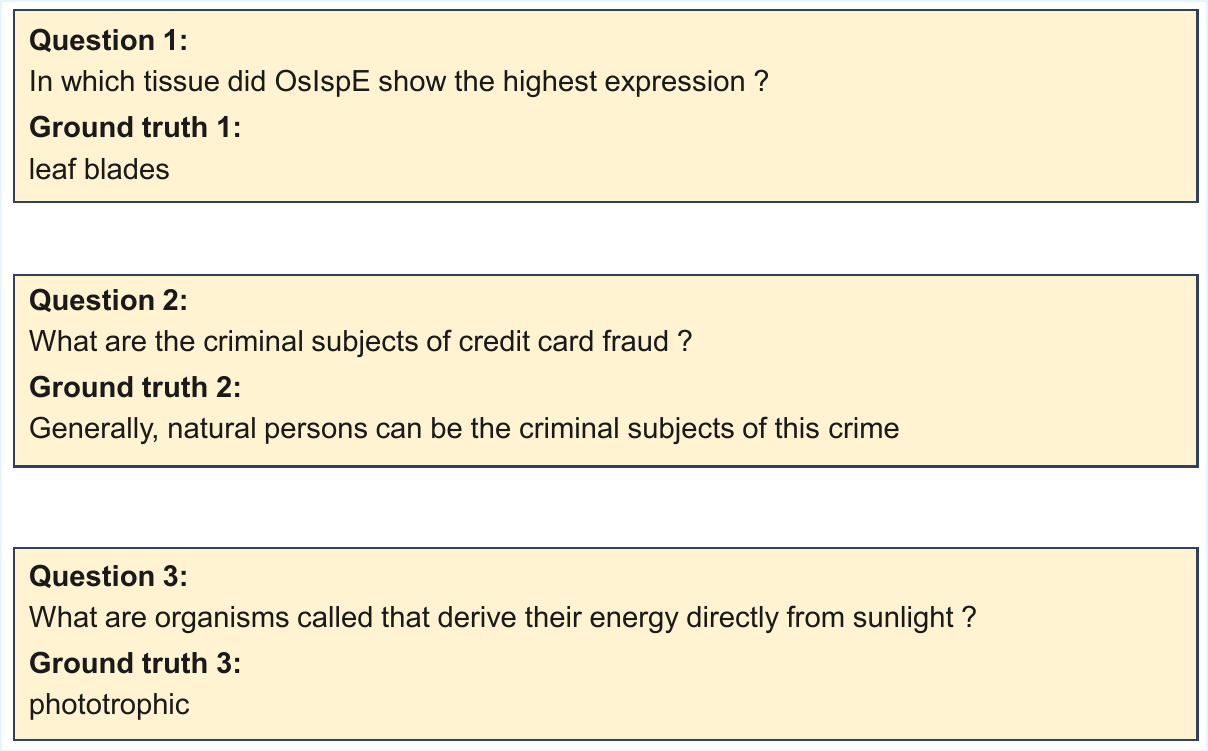}
    \end{center}
    \caption{Three cases of Genie.}
    \label{fig:case_study_genie}
\end{figure*}

As shown in Figure~\ref{fig:case_study_genie}, the data Synthesized by Genie exhibits limitations in both structural complexity and objective verifiability. While the questions address domain-specific topics, they are primarily formatted as simple factual queries that do not provide enough context to support long-chain reasoning. A key drawback is the production of ground truths that are difficult to verify automatically, as exemplified by Question 2. Instead of a concise term or entity, the answer is a descriptive sentence. This conversational and open-ended format poses a challenge for rule-based verifiers, as it is hard to determine correctness through exact matching or simple parsing.

\subsection{Synthesized Data of SDR}
\begin{figure*}[t]
    \begin{center}
        \includegraphics[width=\linewidth]{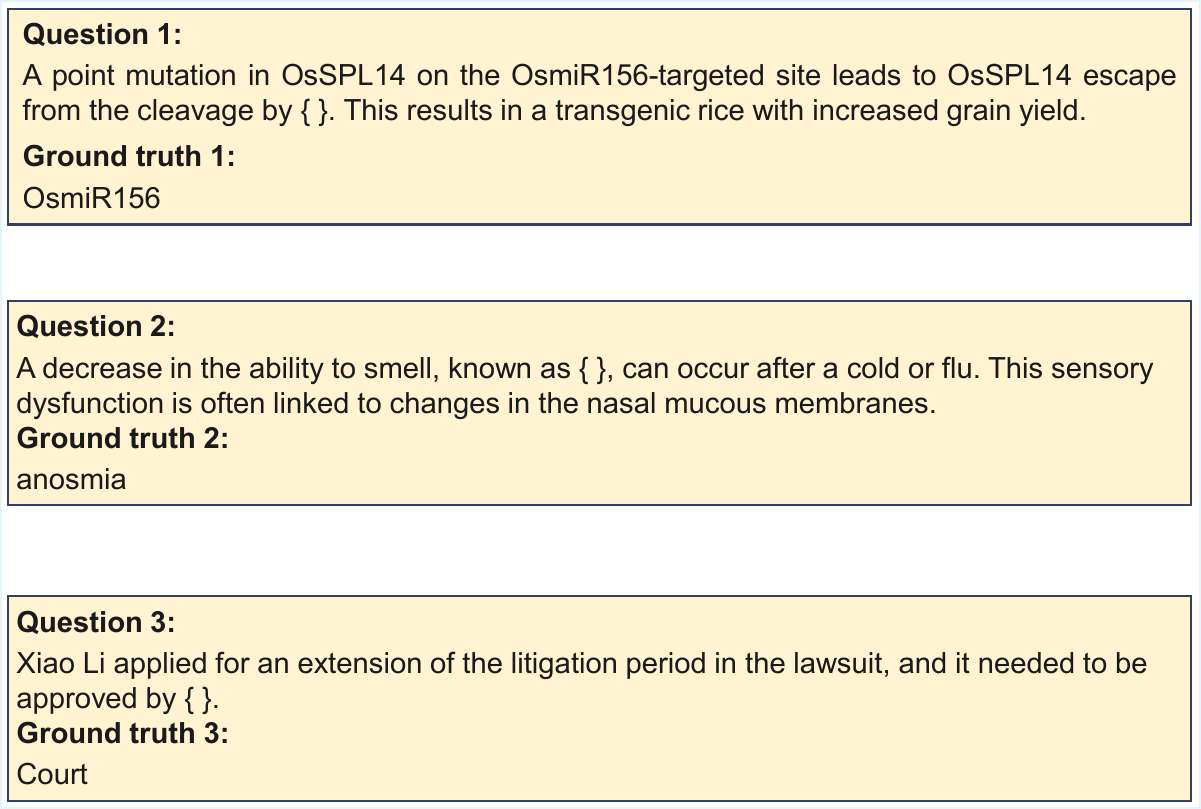}
    \end{center}
    \caption{Three cases of SDR.}
    \label{fig:case_study_SDR}
\end{figure*}

As shown in Figure~\ref{fig:case_study_SDR}, the data format synthesized by SDR can be specified to match the fill-in blank style of K2V. By embedding specific terms within detailed background descriptions, such as biological mechanisms or legal procedures, SDR provides the model with clear logical clues. This setup requires the model to derive the correct answer by integrating the provided context, which ensures the uniqueness of the ground truth. While these questions primarily focus on single-step logical matching and are relatively straightforward compared to complex multi-step reasoning tasks, their rigorous descriptions and accurate knowledge mapping provide a reliable foundation for training models in specialized domains.

\subsection{Synthesized Data of BDS}
\begin{figure*}[t]
    \begin{center}
        \includegraphics[width=\linewidth]{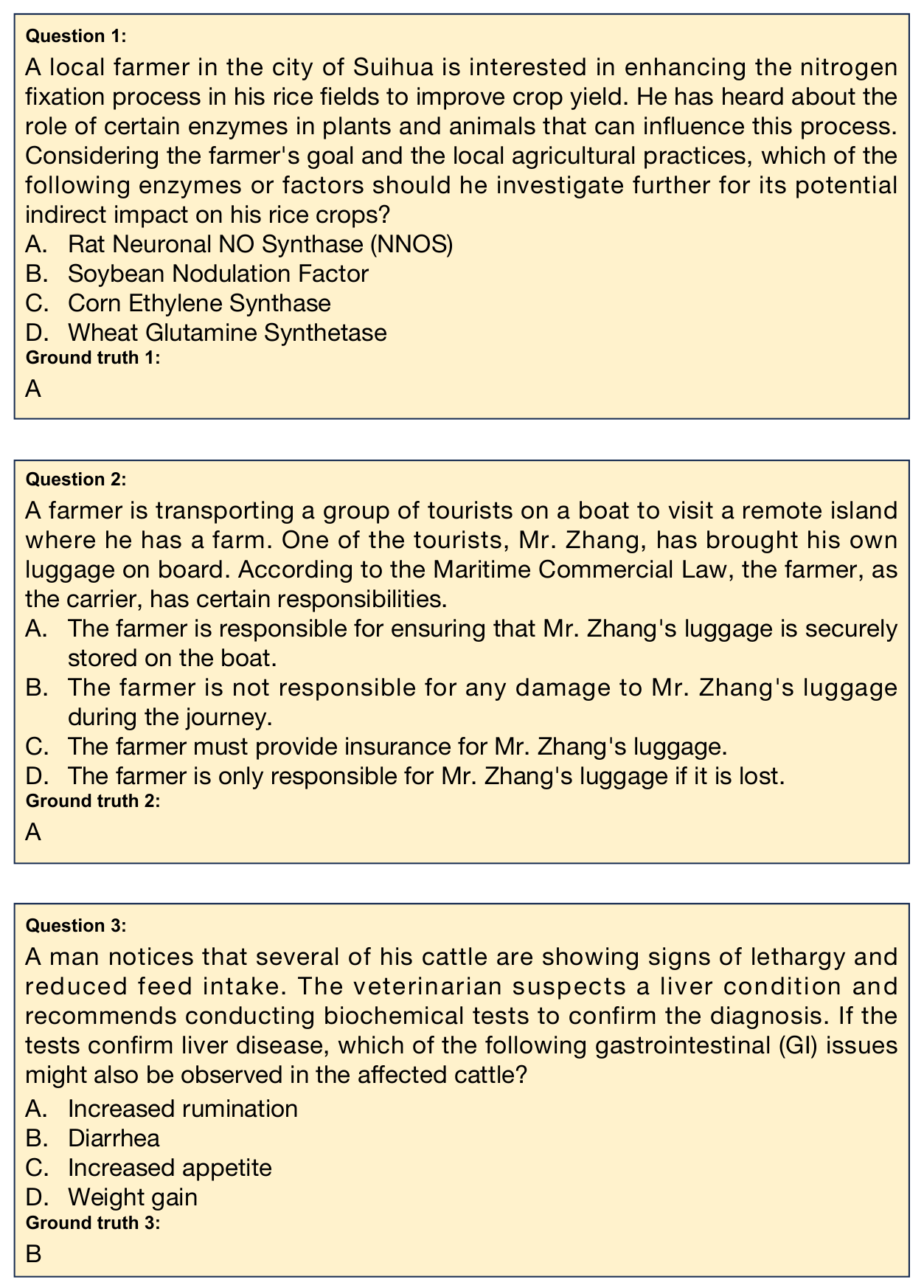}
    \end{center}
    \caption{Three cases of BDS.}
    \label{fig:case_study_BDS}
\end{figure*}

As shown in Figure~\ref{fig:case_study_BDS}, the BDS method adopts a multiple-choice format, which ensures high verifiability for the model. Since answers are restricted to four options, the evaluation system can precisely judge the correctness of outputs, providing clear and unambiguous feedback signals for reinforcement learning. Regarding reasoning depth, this method utilizes knowledge graphs to link fragmented knowledge points into situational scenarios, such as liability determination in maritime law or complication inference for livestock diseases. This requires models to move beyond simple definition retrieval and perform logical integration across multiple knowledge points to make judgments based on complex stems.

\clearpage

\clearpage

\clearpage

\clearpage
\section{Prompts}
\label{Prompts}

K2V utilizes auxiliary LLMs for data synthesis and model training. In this section, we provide the detailed prompts.

\subsection{Prompt of $\mathcal{F}_{\text{text}}$}
\label{Prompt of f_text}
The prompt of the $\mathcal{F}_{\text{text}}$ is as shown in Table~\ref{tab:textualization_function}. This prompt is used to instructs an LLM to convert the masked quintuple into a fill-blank style QA pair.

\subsection{Prompt for NER and RE}
\label{Prompt for NER and RE}
The prompt for NER and RE is as shown in Table~\ref{tab:NER_and_ER}. This prompt is used to instruct an LLM to extract entities and relations from the corpus.

\subsection{Prompt of Judge Model}
\label{Prompt of Judge Model}
The prompt of the Judge Model is as shown in Table~\ref{tab:judge_model}. This prompt is used to instruct an LLM to verify the reasoning process based on a checklist.

\subsection{Prompt for Synthesizing Checklist}
\label{Prompt for Synthesizing Checklist}
The prompt for synthesizing the checklist is as shown in Table~\ref{tab:synthesizing_checklist}.

\input{tex_for_prompts/textualization_function}

\input{tex_for_prompts/NER_and_ER}

\input{tex_for_prompts/judge_model}

\input{tex_for_prompts/synthesizing_checklist}

\clearpage
\section{Examples of General Criteria}
\label{Examples of General Criteria}

K2V synthesizes question-specific checklist based on general criteria, which reflect a high-quality reasoning process and are independent of any specific problem. We develop general criteria for three domains: agriculture (Table~\ref{tab:general_criteria_agri}), medicine (Table~\ref{tab:general_criteria_medical}), and law (Table~\ref{tab:general_criteria_law}). It is worth noting that although different domains require unique general criteria, adapting them to a new domain usually only requires simple keyword substitution  based on an existing one.

\input{general_criteria/agri}

\input{general_criteria/medical}

\input{general_criteria/law}

\clearpage
\section{The Use of LLMs}
\label{The Use of LLMs}

We use OpenAI's GPT-4 as a writing assistant to help improve the clarity, grammar, and style of this manuscript. All scientific ideas, experiments, and conclusions were conceived and executed by the human authors.

\end{document}

%% file: tex_for_prompts/textualization_function.tex
\begin{table*}[!h]
\centering
\begin{minipage}{0.99\linewidth}\vspace{0mm}    \centering
\begin{tcolorbox}[colframe=black!75!white, colback=white, coltitle=white, title=Prompt of the $\mathcal{F}_{\text{text}}$, fonttitle=\bfseries]
\small

\texttt{---}Role\texttt{---}

\vspace{1mm}
You are an NLP expert responsible for generating a logically structured and coherent rephrased version of the TEXT based on ENTITIES and RELATIONSHIPS provided below.
Use English as output language.

\vspace{2mm}
\texttt{---}Goal\texttt{---}

\vspace{1mm}
To generate a version of the text that is rephrased and conveys the same meaning as the original entity and relationship descriptions, while:
\begin{enumerate}[topsep=1pt, itemsep=0.5pt, partopsep=0pt, parsep=0pt, leftmargin=2em]
    \item Following a clear logical flow and structure
    \item Establishing proper cause-and-effect relationships
    \item Ensuring temporal and sequential consistency
    \item Creating smooth transitions between ideas using conjunctions and appropriate linking words like `firstly,' `however,' `therefore,' etc.
\end{enumerate}

\vspace{2mm}
\texttt{---}Instructions\texttt{---}

\vspace{1mm}
1. Analyze the provided ENTITIES and RELATIONSHIPS carefully to identify:
\begin{enumerate}[topsep=1pt, itemsep=0.5pt, partopsep=0pt, parsep=0pt, leftmargin=2em, label={-}]
    \item Key concepts and their hierarchies
    \item Temporal sequences and chronological order
    \item Cause-and-effect relationships
    \item Dependencies between different elements
\end{enumerate}

2. Organize the information in a logical sequence by:
\begin{enumerate}[topsep=1pt, itemsep=0.5pt, partopsep=0pt, parsep=0pt, leftmargin=2em, label={-}]
    \item Starting with foundational concepts
    \item Building up to more complex relationships
    \item Grouping related ideas together
    \item Creating clear transitions between sections
\end{enumerate}

3. Rephrase the text while maintaining:
\begin{enumerate}[topsep=1pt, itemsep=0.5pt, partopsep=0pt, parsep=0pt, leftmargin=2em, label={-}]
    \item Logical flow and progression
    \item Clear connections between ideas
    \item Proper context and background
    \item Coherent narrative structure
\end{enumerate}

4. Review and refine the text to ensure:
\begin{enumerate}[topsep=1pt, itemsep=0.5pt, partopsep=0pt, parsep=0pt, leftmargin=2em, label={-}]
    \item Logical consistency throughout
    \item Clear cause-and-effect relationships
\end{enumerate}

\vspace{2mm}
\hashline

-ENTITIES-

\hashline

\vspace{1mm}
\textit{\{entities\}}

\vspace{2mm}
\hashline

-RELATIONSHIPS-

\hashline

\vspace{1mm}
\textit{\{relationships\}}

\end{tcolorbox}
\caption{Prompt of the $\mathcal{F}_{\text{text}}$. This prompt is used to instructs an LLM to convert the masked quintuple into a fill-blank style QA pair.}
\label{tab:textualization_function}
\end{minipage}
\end{table*}

%% file: tex_for_prompts/NER_and_ER.tex
\begin{table*}[t]
\centering
\begin{minipage}{0.99\linewidth}\vspace{0mm}    \centering
\begin{tcolorbox}[colframe=black!75!white, colback=white, coltitle=white, title=Prompt for NER and RE, fonttitle=\bfseries]
\small

\vspace{1mm}
You are an NLP expert, skilled at analyzing text to extract named entities and their relationships.

\vspace{2mm}
\texttt{---}Goal\texttt{---}

\vspace{1mm}
Given a text document that is potentially relevant to this activity and a list of entity types, identify all entities of those types from the text and all relationships among the identified entities.
Use English as output language.

\vspace{2mm}
\texttt{---}Steps\texttt{---}

\vspace{1mm}
1. Identify all entities. For each identified entity, extract the following information:
\begin{enumerate}[topsep=1pt, itemsep=0.5pt, partopsep=0pt, parsep=0pt, leftmargin=2em, label={-}]
    \item entity\_name: Name of the entity, use same language as input text. If English, capitalized the name.
    \item entity\_type: One of the following types: {concept, date, location, keyword, organization, person, event, work, nature, artificial, science, technology, mission, gene}
    \item entity\_summary: Comprehensive summary of the entity's attributes and activities
    \item Format each entity as: 
    
        \texttt{("entity"<|><entity\_name><|><entity\_type><|><entity\_summary>)}
\end{enumerate}

\vspace{1mm}
2. From the entities identified in step 1, identify all pairs of (source\_entity, target\_entity) that are *clearly related* to each other.
For each pair of related entities, extract the following information:
\begin{enumerate}[topsep=1pt, itemsep=0.5pt, partopsep=0pt, parsep=0pt, leftmargin=2em, label={-}]
    \item source\_entity: name of the source entity, as identified in step 1
    \item target\_entity: name of the target entity, as identified in step 1
    \item relationship\_summary: explanation as to why you think the source entity and the target entity are related to each other
    \item Format each relationship as: 

        \texttt{("relationship"<|><source\_entity><|><target\_entity><|>}
        \texttt{<relationship\_summary>)}
\end{enumerate}

\vspace{1mm}
3. Identify high-level key words that summarize the main concepts, themes, or topics of the entire 

\hspace{3mm} text. These should capture the overarching ideas present in the document.Format the content-level 

\hspace{3mm} key words as \texttt{("content\_keywords"<|><high\_level\_keywords>)}

\vspace{1mm}
4. Return output in Englist as a single list of all the entities and relationships identified. Use **\#\#** 

\hspace{3mm} as the list delimiter.

\vspace{1mm}
5. When finished, output \texttt{<|COMPLETE|>}

\vspace{2mm}
\hashline

-Input Text-

\hashline

\vspace{1mm}
\textit{\{input\_text\}}

\end{tcolorbox}
\caption{Prompt for NER and RE. This prompt is used to instruct an LLM to extract entities and relations from the corpus.
}
\label{tab:NER_and_ER}
\end{minipage}
\end{table*}

%% file: tex_for_prompts/judge_model.tex
\begin{table*}[t]
\centering
\begin{minipage}{0.99\linewidth}\vspace{0mm}    \centering
\begin{tcolorbox}[colframe=black!75!white, colback=white, coltitle=white, title=Prompt of the Judge Model, fonttitle=\bfseries]
\small

\vspace{1mm}
You are an impartial and meticulous AI examiner.

\vspace{1mm}
Your task is to evaluate a student's [Reasoning Process] for a given [Question-Answer Pair] against a specific, detailed [Criterion]. 

\vspace{1mm}
The [Question-Answer Pair] is a fill-in-the-blank question, with \texttt{"\{ \}"} indicating the content to be filled in. A fill-in-the-blank question may contain multiple \texttt{"\{ \}"}, and the content to be filled in for each \texttt{"\{ \}"} is the same. 

\vspace{1mm}
Your judgment must be strict, objective, and based solely on the provided information.

\vspace{2mm}
NOTE: Your output can only be \texttt{"yes"} or \texttt{"NO"}

\vspace{2mm}
[Question-Answer Pair]

question: \textit{question}

answer: \textit{answer}

\vspace{2mm}
[Criterion]

criterion: \textit{criterion}

\vspace{2mm}
[Reasoning Process]

reasoning process: \textit{reasoning process}

\end{tcolorbox}
\caption{Prompt of the Judge Model. This prompt is used to instruct an LLM to verify the reasoning process based on a checklist.}
\label{tab:judge_model}
\end{minipage}
\end{table*}

%% file: tex_for_prompts/synthesizing_checklist.tex
\begin{table*}[t]
\centering
\begin{minipage}{0.99\linewidth}\vspace{0mm}    \centering
\begin{tcolorbox}[colframe=black!75!white, colback=white, coltitle=white, title=Prompt for synthesizing the checklist, fonttitle=\bfseries]
\small

\vspace{1mm}
You are a senior expert in agriculture and biology, specializing in creating and grading exam questions. Your task is to create a set of detailed scoring checklist for a [Specific Question] based on the provided [General Criteria].

\vspace{2mm}
[Specific Question]:

\vspace{1mm}
A complete question in the field of agriculture and biology, including the question and the corresponding answer.

\vspace{2mm}
[General Criteria]:

\vspace{1mm}
Concepts and Knowledge:
\begin{enumerate}[topsep=1pt, itemsep=0.5pt, partopsep=0pt, parsep=0pt, leftmargin=2em]
    \item Accurately defines the core biological concepts involved in the question.
    \item Clearly describes the involved biological processes in the correct logical sequence.
    \item Accurately explains the meaning and relationships represented by abstract biological models in words.
    \item Applies abstract biological concepts to the given specific scenario.
    \item Correctly explains the connection between a biological concept or process and other related principles.
\end{enumerate}

\vspace{1mm}
Scientific Method and Design:
\begin{enumerate}[topsep=1pt, itemsep=0.5pt, partopsep=0pt, parsep=0pt, leftmargin=2em]
    \item Clearly states a relevant null hypothesis or alternative hypothesis.
    \item Accurately identifies the independent, dependent, and key control variables of an experiment.
    \item Makes a logical and reasonable prediction of the experimental outcome based on a scientific hypothesis.
    \item Evaluates the validity or potential flaws of a given experimental design.
\end{enumerate}

\vspace{1mm}
Statistics and Evaluation:
\begin{enumerate}[topsep=1pt, itemsep=0.5pt, partopsep=0pt, parsep=0pt, leftmargin=2em]
    \item In appropriate contexts, correctly uses statistical concepts to explain the reliability of data.
    \item Based on data analysis, draws a conclusion of "support," "refute," or "inconclusive" for a given scientific hypothesis.
    \item Explains outliers or anomalous data points and analyzes their potential causes or impact on the conclusion.
\end{enumerate}

\vspace{1mm}
Argumentation and Reasoning:
\begin{enumerate}[topsep=1pt, itemsep=0.5pt, partopsep=0pt, parsep=0pt, leftmargin=2em]
    \item Makes a scientific claim that is specific and supported by concrete evidence.
    \item Clearly articulates how the evidence supports the scientific claim, demonstrating a strong logical chain.
    \item Predicts the likely consequences of a change (e.g., disturbance, mutation) to a system based on biological principles.
    \item Explains the underlying biological reason for an observed phenomenon or experimental result.
    \item Avoids over-extrapolation or unfounded speculation beyond the scope of the given evidence.
    \item The overall response is well-structured, logically coherent, and clearly written, avoiding self-contradictions and redundant statements.
\end{enumerate}

\vspace{2mm}
Based on the [General Criteria] above, design a set of detailed and objectively scorable checklist for the provided [Specific Exam Question]. The checklist will be used to evaluate the student's problem-solving approach (reasoning process).
The checklist should consist of multiple independent criteria. Each criteria must be a clear, specific statement describing what an ideal step or thought process should achieve, making it objectively assessable. Please ensure The checklist are closely related to the core knowledge and skill requirements of the [Specific Exam Question].
Only output the checklist, with no other content. Please structure the output in JSON format. For example:

\texttt{["criteria 1", "criteria 2",]}

\end{tcolorbox}
\caption{Prompt for synthesizing the checklist.}
\label{tab:synthesizing_checklist}
\end{minipage}
\end{table*}

%% file: general_criteria/agri.tex
\begin{table*}[t]
\centering
\begin{minipage}{0.99\linewidth}\vspace{0mm}    \centering
\begin{tcolorbox}[
    colback=orange!5!white,      
    colframe=orange!75!black,    
    coltitle=white,              
    title=General criteria in the agricultural domain, 
    fonttitle=\bfseries
]
\small

\vspace{1mm}
Concepts and Knowledge:
\begin{enumerate}[topsep=1pt, itemsep=0.5pt, partopsep=0pt, parsep=0pt, leftmargin=2em]
    \item Accurately defines the core biological concepts involved in the question.
    \item Clearly describes the involved biological processes in the correct logical sequence.
    \item Accurately explains the meaning and relationships represented by abstract biological models in words.
    \item Applies abstract biological concepts to the given specific scenario.
    \item Correctly explains the connection between a biological concept or process and other related principles.
\end{enumerate}

\vspace{1mm}
Scientific Method and Design:
\begin{enumerate}[topsep=1pt, itemsep=0.5pt, partopsep=0pt, parsep=0pt, leftmargin=2em]
    \item Clearly states a relevant null hypothesis or alternative hypothesis.
    \item Accurately identifies the independent, dependent, and key control variables of an experiment.
    \item Makes a logical and reasonable prediction of the experimental outcome based on a scientific hypothesis.
    \item Evaluates the validity or potential flaws of a given experimental design.
\end{enumerate}

\vspace{1mm}
Data Processing and Analysis:
\begin{enumerate}[topsep=1pt, itemsep=0.5pt, partopsep=0pt, parsep=0pt, leftmargin=2em]
    \item Accurately and correctly extracts key data points.
    \item Clearly and comprehensively describes the overall trend or significant patterns in the given data.
    \item Accurately describes the relationship between variables (e.g., positive correlation, negative correlation, no correlation).
    \item Correctly performs necessary mathematical calculations (e.g., rate, rate of change, percentage) to support the analysis.
\end{enumerate}

\vspace{1mm}
Statistics and Evaluation:
\begin{enumerate}[topsep=1pt, itemsep=0.5pt, partopsep=0pt, parsep=0pt, leftmargin=2em]
    \item In appropriate contexts, correctly uses statistical concepts to explain the reliability of data.
    \item Based on data analysis, draws a conclusion of "support," "refute," or "inconclusive" for a given scientific hypothesis.
    \item Explains outliers or anomalous data points and analyzes their potential causes or impact on the conclusion.
\end{enumerate}

\vspace{1mm}
Argumentation and Reasoning:
\begin{enumerate}[topsep=1pt, itemsep=0.5pt, partopsep=0pt, parsep=0pt, leftmargin=2em]
    \item Makes a scientific claim that is specific and supported by concrete evidence.
    \item Clearly articulates how the evidence supports the scientific claim, demonstrating a strong logical chain.
    \item Predicts the likely consequences of a change (e.g., disturbance, mutation) to a system based on biological principles.
    \item Explains the underlying biological reason for an observed phenomenon or experimental result.
    \item Avoids over-extrapolation or unfounded speculation beyond the scope of the given evidence.
    \item The overall response is well-structured, logically coherent, and clearly written, avoiding self-contradictions and redundant statements.
\end{enumerate}

\end{tcolorbox}
\caption{General criteria in the agricultural domain.}
\label{tab:general_criteria_agri}
\end{minipage}
\end{table*}

%% file: general_criteria/medical.tex
\begin{table*}[t]
\centering
\begin{minipage}{0.99\linewidth}\vspace{0mm}    \centering
\begin{tcolorbox}[
    colback=orange!5!white,      
    colframe=orange!75!black,    
    coltitle=white,              
    title=General criteria in the medical domain, 
    fonttitle=\bfseries
]
\small

\vspace{1mm}
Concepts and Knowledge:
\begin{enumerate}[topsep=1pt, itemsep=0.5pt, partopsep=0pt, parsep=0pt, leftmargin=2em]
    \item Accurately defines the core medical concepts involved in the question.
    \item Clearly describes the involved medical processes in the correct logical sequence.
    \item Accurately explains the meaning and relationships represented by abstract biological or medical models in words.
    \item Applies abstract biological or medical concepts to the given specific scenario.
    \item Correctly explains the connection between a medical concept or process and other related principles.
\end{enumerate}

\vspace{1mm}
Scientific Method and Design:
\begin{enumerate}[topsep=1pt, itemsep=0.5pt, partopsep=0pt, parsep=0pt, leftmargin=2em]
    \item Clearly states a relevant null hypothesis or alternative hypothesis.
    \item Accurately identifies the independent, dependent, and key control variables of an experiment.
    \item Makes a logical and reasonable prediction of the experimental outcome based on a scientific hypothesis.
    \item Evaluates the validity or potential flaws of a given experimental design.
\end{enumerate}

\vspace{1mm}
Data Processing and Analysis:
\begin{enumerate}[topsep=1pt, itemsep=0.5pt, partopsep=0pt, parsep=0pt, leftmargin=2em]
    \item Accurately and correctly extracts key data points.
    \item Clearly and comprehensively describes the overall trend or significant patterns in the given data.
    \item Accurately describes the relationship between variables (e.g., positive correlation, negative correlation, no correlation).
    \item Correctly performs necessary mathematical calculations (e.g., rate, rate of change, percentage) to support the analysis.
\end{enumerate}

\vspace{1mm}
Statistics and Evaluation:
\begin{enumerate}[topsep=1pt, itemsep=0.5pt, partopsep=0pt, parsep=0pt, leftmargin=2em]
    \item In appropriate contexts, correctly uses statistical concepts to explain the reliability of data.
    \item Based on data analysis, draws a conclusion of "support," "refute," or "inconclusive" for a given scientific hypothesis.
    \item Explains outliers or anomalous data points and analyzes their potential causes or impact on the conclusion.
\end{enumerate}

\vspace{1mm}
Argumentation and Reasoning:
\begin{enumerate}[topsep=1pt, itemsep=0.5pt, partopsep=0pt, parsep=0pt, leftmargin=2em]
    \item Makes a scientific claim that is specific and supported by concrete evidence.
    \item Clearly articulates how the evidence supports the scientific claim, demonstrating a strong logical chain.
    \item Predicts the likely consequences of a change (e.g., disturbance, mutation) to a system based on biological or medical principles.
    \item Explains the underlying biological or medical reason for an observed phenomenon or experimental result.
    \item Avoids over-extrapolation or unfounded speculation beyond the scope of the given evidence.
    \item Based on diagnostic or analytical results, proposes specific and feasible treatment or management recommendations that comply with clinical guidelines and ethical principles.
    \item Clearly articulates the rationale for the proposed recommendations and weighs their potential benefits and risks.
    \item Be able to ignore irrelevant information and focus on answering the question directly.
    \item The overall response is well-structured, logically coherent, and clearly written, avoiding self-contradictions and redundant statements.
\end{enumerate}

\end{tcolorbox}
\caption{General criteria in the medical domain}
\label{tab:general_criteria_medical}
\end{minipage}
\end{table*}

%% file: general_criteria/law.tex
\begin{table*}[t]
\centering
\begin{minipage}{0.99\linewidth}\vspace{0mm}    \centering
\begin{tcolorbox}[
    colback=orange!5!white,      
    colframe=orange!75!black,    
    coltitle=white,              
    title=General criteria in the legal domain, 
    fonttitle=\bfseries
]
\small

\vspace{1mm}
Fact and Issue Identification:
\begin{enumerate}[topsep=1pt, itemsep=0.5pt, partopsep=0pt, parsep=0pt, leftmargin=2em]
    \item Accurately identifies and extracts key legally relevant facts from the case material.
    \item Clearly and accurately identifies the core legal issues or points of contention presented in the case.
    \item Is able to distinguish between legally relevant and irrelevant facts.
II. Rule Statement and Interpretation
\end{enumerate}

\vspace{1mm}
Rule Statement and Interpretation:
\begin{enumerate}[topsep=1pt, itemsep=0.5pt, partopsep=0pt, parsep=0pt, leftmargin=2em]
    \item Accurately states the legal rules (including statutes, judicial interpretations, fundamental principles, etc.) relevant to the identified issues.
    \item Correctly explains the meaning and constituent elements of the legal rules.
    \item Where appropriate, is able to articulate the legislative intent, value orientation, or legal theory behind the relevant rules.
\end{enumerate}

\vspace{1mm}
Application and Analysis:
\begin{enumerate}[topsep=1pt, itemsep=0.5pt, partopsep=0pt, parsep=0pt, leftmargin=2em]
    \item Effectively connects the identified key facts to the relevant legal rules (i.e., the process of "subsumption").
    \item Logically analyzes whether the facts of the case satisfy (or fail to satisfy) the constituent elements of the legal rules.
    \item Is able to analyze and argue from the perspectives of all involved parties (e.g., plaintiff/defendant, prosecution/defense).
    \item Is able to anticipate and respond to potential and compelling counterarguments or defenses.
    \item When dealing with complex problems, is able to conduct a layered, step-by-step analysis of different claims or legal relationships.
\end{enumerate}

\vspace{1mm}
Conclusion and Consequences:
\begin{enumerate}[topsep=1pt, itemsep=0.5pt, partopsep=0pt, parsep=0pt, leftmargin=2em]
    \item Based on the preceding analysis, draws a clear, reasonable, and persuasive conclusion for each issue.
    \item Is able to articulate the specific legal consequences corresponding to the conclusion (e.g., type and scope of civil liability, determination of criminal responsibility).
    \item Proposes solutions or legal advice that are specific, feasible, and in compliance with legal regulations and professional ethics.
\end{enumerate}

\vspace{1mm}
Overall Structure and Expression:
\begin{enumerate}[topsep=1pt, itemsep=0.5pt, partopsep=0pt, parsep=0pt, leftmargin=2em]
    \item The overall response is clearly structured and logically coherent (e.g., follows a framework like IRAC: Issue, Rule, Application, Conclusion).
    \item Uses legal terminology accurately and appropriately.
    \item The reasoning process is rigorous, avoiding over-extrapolation or speculation not supported by the given facts or law.
    \item Is able to ignore irrelevant information and focus on answering the question directly.
    \item The overall response is well-written, clear, and avoids self-contradictions or unnecessary redundancy.
    \item The overall response is clearly written, avoiding self-contradictions and redundant statements.
\end{enumerate}

\end{tcolorbox}
\caption{General criteria in the legal domain}
\label{tab:general_criteria_law}
\end{minipage}
\end{table*}